\documentclass[a4paper]{cas-dc}

\usepackage{enumitem}
\usepackage{placeins} 

\ExplSyntaxOn
\cs_set:Npn \__first_footerline: {}
\ExplSyntaxOff

\usepackage[authoryear]{natbib}

\renewcommand{\topfraction}{0.95}
\renewcommand{\dbltopfraction}{0.95}
\renewcommand{\textfraction}{0.05}
\renewcommand{\floatpagefraction}{0.80}
\renewcommand{\dblfloatpagefraction}{0.80}


\def\tsc#1{\csdef{#1}{\textsc{\lowercase{#1}}\xspace}}
\tsc{WGM}
\tsc{QE}

\begin{document}
\let\WriteBookmarks\relax
\def\floatpagepagefraction{1}
\def\textpagefraction{.001}
\let\printorcid\relax

\setcounter{topnumber}{4}
\setcounter{dbltopnumber}{4}
\renewcommand{\topfraction}{0.95}
\renewcommand{\dbltopfraction}{0.95}
\renewcommand{\textfraction}{0.05}
\renewcommand{\floatpagefraction}{0.80}
\renewcommand{\dblfloatpagefraction}{0.95}

\shorttitle{}    

\shortauthors{}  

\title [mode = title]{EchoDiST: Self-distillation-based joint learning for diffusion-conditioned echocardiographic myocardial motion estimation}  



%

\author[1]{Feiyue Qi}
\credit{Writing – review \& editing, Writing – original draft, Visualization, Software, Methodology, Formal analysis, Data curation, Conceptualization}
\fnmark[1]

\author[1]{Xingyue Wei}
\credit{Writing – review \& editing, Methodology, Data curation, Investigation, Formal analysis}
\fnmark[1]

\author[1]{Jianwen Luo}
\cormark[1]
\ead{luo_jianwen@tsinghua.edu.cn}
\credit{Writing – review \& editing, Funding acquisition, Methodology, Supervision, Resources}

\cortext[1]{Corresponding author}

\fntext[1]{These authors contributed equally to this work.}

\affiliation[1]{
  organization={School of Biomedical Engineering, Tsinghua Medicine, Tsinghua University},
  city={Beijing},
  postcode={100084},
  country={China}
}


\begin{abstract}
Motion estimation in echocardiography is essential for quantitative assessment of cardiac function and myocardial mechanics. However, accurate myocardial motion estimation remains challenging due to image artifacts, limited image information, speckle decorrelation, and the scarcity of ground-truth myocardial displacement fields. Anatomy-guided approaches can provide additional structural information for motion estimation, but often depend on expert-labeled myocardial segmentations, which limits their scalability in clinical applications. To address these challenges, we propose EchoDiST, a framework for unsupervised echocardiographic myocardial motion estimation that integrates a self-distillation-based joint learning strategy with a diffusion-conditioned motion estimation network. Here, unsupervised motion estimation refers to learning without ground-truth displacement fields. The self-distillation-based joint learning strategy enables anatomical segmentation and myocardial motion estimation to be jointly optimized under limited anatomical annotations. The diffusion-conditioned motion estimation network incorporates diffusion-based conditioning into the estimation process, with stochastic diffusion perturbations used only during training, while inference is performed through a single deterministic forward pass without iterative reverse-diffusion sampling. We extensively evaluated EchoDiST on three echocardiographic datasets, including two external test datasets evaluated under cross-view and cross-dataset settings. Compared with seven representative learning-based methods, EchoDiST demonstrates consistent improvements in myocardial motion estimation across multiple evaluation levels, including anatomical alignment, myocardial strain assessment, and motion-derived functional and cardiac-phase assessment. These gains are statistically significant, further supporting the effectiveness and robustness of EchoDiST across the evaluated tasks and datasets. Overall, EchoDiST provides an effective approach for reliable myocardial motion estimation under limited anatomical supervision and supports downstream quantitative assessment of cardiac function.
\end{abstract}


\maketitle

\section{Introduction}\label{sec:introduction}
Cardiovascular diseases (CVDs) remain the leading cause of death worldwide and continue to impose substantial economic and societal burdens~\citep{Tsao2022}. Quantitative assessment of myocardial motion can reveal regional and global mechanical abnormalities, thereby providing clinically useful information for disease detection, functional assessment, prognostic evaluation, and prediction of therapeutic response~\citep{Risum2012,Park2018,Zhang2018,Van2019}. Echocardiography is particularly well suited to myocardial motion assessment because it is non-invasive, widely available and cost-effective, while also providing information on both global cardiac function and regional myocardial mechanics~\citep{Ahn2023}. Reliable myocardial motion estimation from echocardiographic sequences is therefore essential for the quantitative assessment of cardiac function~\citep{Ta2024}.

Despite its importance, accurate myocardial motion estimation from echocardiographic sequences remains challenging. Echocardiographic images are inherently affected by artifacts (e.g., shadowing and reverberation), limited image information, and speckle decorrelation. These factors can alter the appearance of the same myocardial tissue across cardiac phases, making reliable motion estimation difficult, particularly during rapid or rotational motion~\citep{Evain2022}. Meanwhile, the scarcity of ground-truth myocardial displacement fields in clinical echocardiography limits the feasibility of fully supervised motion estimation~\citep{Ahn2023}.

To address these challenges, anatomical or shape information, particularly myocardial segmentation, has been incorporated into cardiac motion estimation to improve its accuracy~\citep{Papademetris2002,Ta2021,Ta2024}. However, inaccurate segmentation may in turn lead to unreliable motion estimation~\citep{Ta2024}. Previous studies have further explored joint optimization of segmentation and motion estimation, allowing the two tasks to benefit from each other ~\citep{Qin2018,Ta2020,Ta2024}. However, existing anatomy-guided motion estimation approaches often rely on expert-labeled myocardial segmentations to provide structural supervision~\citep{Ta2021,Ahn2023}. Obtaining such manual annotations at scale is difficult, labor-intensive, and time-consuming~\citep{Xu2019,Ta2021}. Reducing this dependence while maintaining effective joint learning therefore remains a key challenge.

In this work, we propose EchoDiST (Echocardiographic Diffusion-conditioned Self-distillation), a framework for unsupervised echocardiographic myocardial motion estimation that integrates a self-distillation-based joint learning strategy with a diffusion-conditioned motion estimation network. Here, unsupervised motion estimation refers to learning myocardial motion without ground-truth displacement fields. The self-distillation-based joint learning strategy enables anatomical segmentation and myocardial motion estimation to be jointly optimized under limited anatomical annotations.  The diffusion-conditioned motion estimation network incorporates diffusion-based conditioning into the estimation process, providing an alternative to conventional direct regression-based motion estimation. To the best of our knowledge, diffusion-conditioned motion estimation has not previously been explored for myocardial motion estimation. Comprehensive experiments on three echocardiographic datasets demonstrate the effectiveness of EchoDiST in myocardial motion estimation across multiple evaluation levels, including anatomical alignment, myocardial strain assessment, and motion-derived functional and cardiac-phase assessment.

The main contributions of this work are summarized as follows:

\begin{itemize}
    \setlength{\itemsep}{2pt}
    \setlength{\parskip}{0pt}
    \setlength{\parsep}{0pt}
    \item We propose EchoDiST for echocardiographic myocardial motion estimation. The framework incorporates diffusion-based conditioning into motion estimation and integrates anatomical segmentation and motion estimation to improve reliability.

    \item We develop a self-distillation-based joint learning strategy that simultaneously optimizes anatomical segmentation and unsupervised myocardial motion estimation under limited anatomical annotations. The proposed strategy uses the segmentation results and estimated motion to provide mutual learning signals, thereby reducing the reliance on extensive manual segmentation annotations.

    \item We conduct a comprehensive evaluation of EchoDiST across three echocardiographic datasets, including two external test datasets. The results demonstrate consistent improvements in myocardial motion estimation across multiple evaluation levels, including anatomical alignment, myocardial strain assessment, and motion-derived functional and cardiac-phase assessment.
\end{itemize}

\section{Related Work}\label{sec:related work}
\subsection{Echocardiographic Motion Estimation Methods}

Echocardiographic motion estimation aims to estimate myocardial displacement across cardiac phases, providing a basis for subsequent cardiac functional analysis. Conventional approaches primarily depend on intensity- or feature-based tracking. \citet{Huang2014} employed sparse representation and dictionary learning for echocardiographic contour tracking, while \citet{Parajuli2016} integrated dynamic shape tracking with radio-frequency speckle tracking for cardiac motion analysis. Despite these advances, conventional motion-tracking approaches can be computationally intensive and often rely on prior knowledge or extensively hand-crafted features~\citep{Ta2024}.

Deep learning-based methods have subsequently been developed to estimate myocardial motion directly from echocardiographic data. Some supervised approaches rely on synthetic data with reference displacement fields; for example, \citet{Evain2022} developed a deep-learning framework for 2D echocardiographic motion estimation using realistic synthetic B-mode sequences with reference myocardial motion. However, reliable dense motion annotations are difficult to obtain from clinical echocardiography, motivating unsupervised approaches that estimate motion without direct supervision from ground-truth displacement fields. \citet{Ahn2023}, for instance, proposed a co-attention spatial transformer network for unsupervised motion tracking in 3D echocardiography by leveraging inter-frame spatiotemporal information. Although such approaches alleviate the need for motion annotations, their optimization is mainly driven by image appearance or spatiotemporal correspondence, with limited explicit anatomical supervision.

Anatomical labels have therefore been introduced as auxiliary supervision for echocardiographic motion estimation. \citet{Ta2021}, for instance, used manually traced segmentations to provide shape guidance for left-ventricular motion estimation in 3D+Time echocardiography. Furthermore, \citet{Ta2024} showed that anatomical segmentation and motion estimation can be complementary when jointly optimized through a multi-task learning framework. However, extending such supervision to large-scale motion estimation would require a large number of expert-annotated ground-truth segmentation masks, which are costly and labor-intensive to obtain. 

To reduce this dependence, our framework learns anatomical segmentations from a limited labeled subset and uses the resulting predictions to provide structural guidance over a larger set of training pairs for motion estimation.

\subsection{Diffusion-Based Motion Estimation Methods}

Diffusion models learn to reverse a progressive noising process through a denoising objective~\citep{Ho2020}. Originally developed for generative modeling, diffusion models have subsequently been adapted to dense motion estimation between medical images, where diffusion-derived features can provide spatial information useful for correspondence learning. DiffuseMorph, for example, models the conditional score associated with the spatial transformation between a moving–fixed image pair and uses the resulting latent feature to generate a dense displacement field~\citep{Kim2022}.

Subsequent studies have explored different ways of incorporating diffusion into motion estimation. FSDiffReg employs multi-scale semantic features derived from a diffusion model to guide dense displacement prediction, while the diffusion score further contributes to topology-preserving optimization~\citep{Qin2023}. In contrast, DiffuseReg directly formulates the displacement field as the diffusion target and progressively recovers it through iterative denoising~\citep{Zhuo2024}. More recently, LDM-Morph incorporates semantic features extracted from a latent diffusion model and integrates them with global image features through cross-attention, together with a hierarchical similarity measure defined in both pixel and latent-feature spaces~\citep{Wu2026}.

Collectively, these studies demonstrate the potential of diffusion-based modeling for dense motion estimation between medical images, yet its application to echocardiographic motion estimation remains relatively underexplored, particularly for diffusion-conditioned learning of myocardial motion. Motivated by this gap, we employ diffusion-conditioned learning for direct echocardiographic motion estimation while retaining deterministic single-pass inference.

\subsection{Joint Learning of Segmentation and Registration with Limited Annotations}

Anatomical guidance is widely used to impose structural constraints on deformable registration, but its reliance on dense segmentation annotations limits its use in label-scarce settings. To reduce the dependence of anatomy-guided registration on densely annotated data, several studies have explored learning anatomical supervision from limited labels and extending it to unlabeled images. DeepAtlas~\citep{Xu2019} jointly learned weakly supervised registration and semi-supervised segmentation, allowing predicted segmentations to provide anatomical guidance when manual annotations were unavailable. \citet{He2022} further introduced a few-shot deformable registration framework that decouples anatomical perception from image correspondence and uses reverse teaching to transfer structural knowledge from only a few labeled images to unlabeled data. \citet{Shi2023} incorporated self-training into joint registration and segmentation, where quality-assessed pseudo-labels extend anatomical supervision to unlabeled images while supporting weakly supervised registration. Alongside these learning strategies, pretrained medical vision models have provided an additional source of anatomical information for registration. \citet{Xu2025} employed text-prompted segmentation masks generated by a vision foundation model to explicitly encode anatomical structures for deformable registration, using the resulting masks for feature fusion, semantic prototype alignment, and contour-aware optimization. 

Inspired by these studies, we propose EchoDiST, which jointly learns anatomical segmentation and unsupervised motion estimation through EMA-based self-distillation. Such a mechanism can  achieve effective learning with limited annotations.

\section{Methods}\label{sec:Method}

\begin{figure*}
    \centering
    \includegraphics[width=\textwidth]{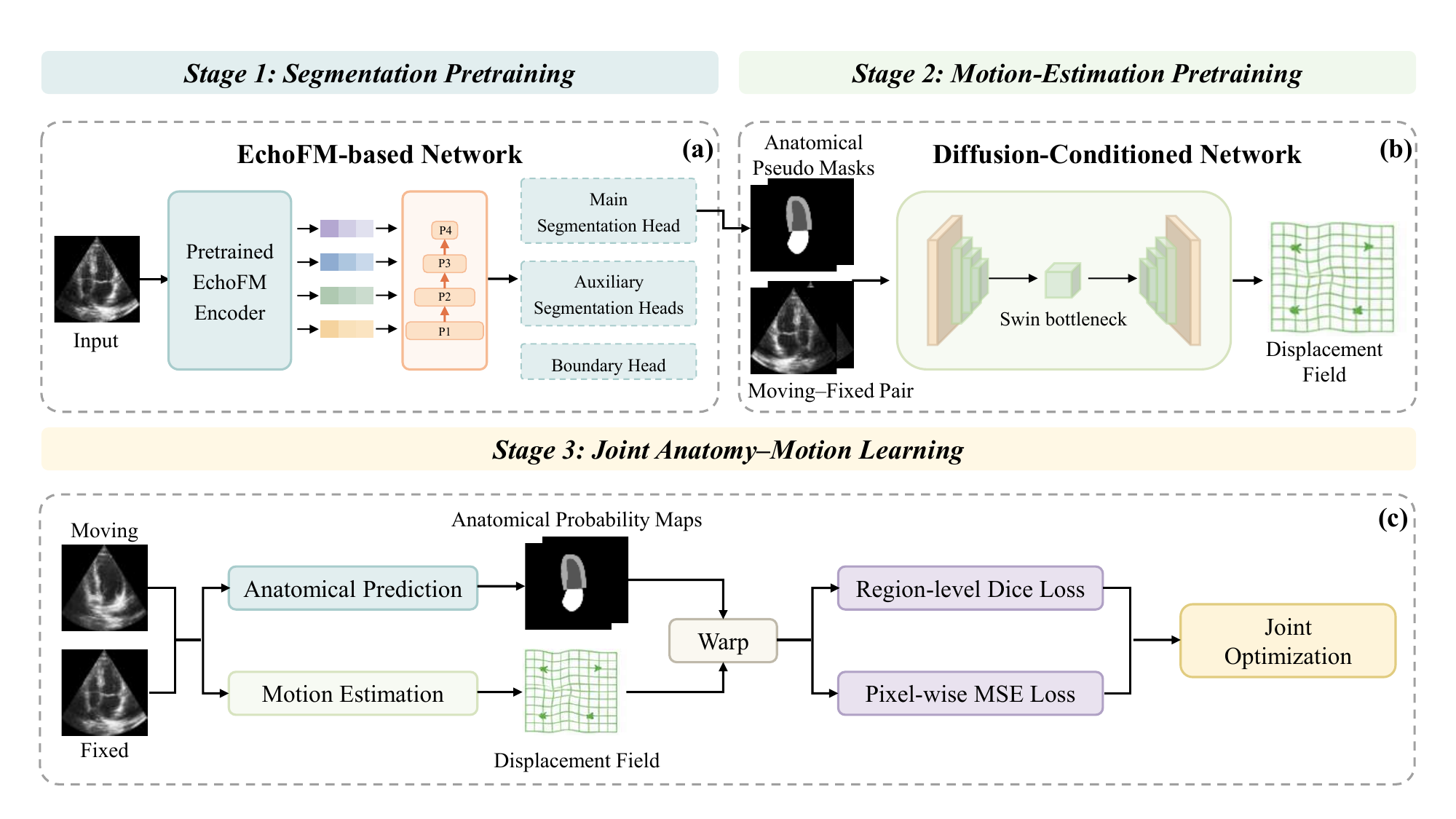}
    \caption{Overview of the proposed EchoDiST framework for echocardiographic motion estimation. (a) Stage 1 pretrains an EchoFM-based segmentation network using a limited set of manually annotated images to generate anatomical pseudo masks. (b) Stage 2 pretrains the diffusion-conditioned motion-estimation network to estimate the displacement field, with anatomical pseudo masks providing structural supervision. (c) Stage 3 jointly optimizes anatomical segmentation and motion estimation through the EMA-based self-distillation strategy.}
    \label{fig:overview}
\end{figure*}

Given a moving echocardiographic frame $I_m$ and a fixed frame $I_f$, our goal is to estimate a dense displacement field $\mathbf{u}_{m\to f}$ that establishes spatial correspondence between the two frames. No ground-truth displacement fields are used for motion-estimation learning. As illustrated in Fig.~\ref{fig:overview}, the proposed EchoDiST follows a three-stage training pipeline:

\begin{enumerate}[
    label=\arabic*.,
    leftmargin=0pt,
    labelindent=0pt,
    labelsep=0.5em,
    itemindent=*
]
    \item \textbf{Segmentation Pretraining Stage:}
    A segmentation network is pretrained on a limited set of annotated echocardiographic images to provide structural guidance for subsequent motion estimation.
    \item \textbf{Motion-Estimation Pretraining Stage:}
    The diffusion-conditioned motion-estimation network is pretrained with the segmentation masks predicted by the segmentation network as anatomical guidance. Gaussian perturbations at randomly sampled diffusion timesteps facilitate motion estimation across different perturbation levels.
    \item \textbf{Self-distillation-based Joint Learning Stage:}
    The pretrained segmentation and motion-estimation networks are jointly optimized through EMA-based self-distillation and reliability-weighted bidirectional cross-frame anatomical consistency.
\end{enumerate}

\subsection{Segmentation Pretraining}

In Stage 1, a segmentation network is pretrained using a limited set of manually annotated echocardiographic images. To enhance the robustness of anatomical segmentation, its encoder adopts the same architecture as EchoFM~\citep{Kim2025}, a foundation model pretrained on a large-scale collection of echocardiographic images. An FPN-based decoder~\citep{Lin2017} integrates multi-level features for anatomical segmentation. The main segmentation head predicts four classes: background, LV cavity, myocardium, and LA. Auxiliary segmentation heads provide deep supervision at intermediate FPN levels, while a lightweight boundary branch encourages boundary-aware learning.

The segmentation-pretraining objective $\mathcal{L}_{\mathrm{seg}}$ comprises the main segmentation loss $\mathcal{L}_{\mathrm{main}}$, auxiliary deep-supervision loss $\mathcal{L}_{\mathrm{aux}}$, and boundary loss $\mathcal{L}_{\mathrm{bnd}}$:

\begin{equation}
\mathcal{L}_{\mathrm{main}}
=
\mathcal{L}_{\mathrm{wce}}(P,Y)
+
\mathcal{L}_{\mathrm{dice}}(P,Y),
\label{eq:seg_main}
\end{equation}

\begin{equation}
\mathcal{L}_{\mathrm{aux}}
=
\frac{1}{S}
\sum_{s=1}^{S}
\left[
\mathcal{L}_{\mathrm{wce}}(P_s^{\mathrm{aux}},Y_s)
+
\mathcal{L}_{\mathrm{dice}}(P_s^{\mathrm{aux}},Y_s)
\right],
\label{eq:seg_aux}
\end{equation}

\begin{equation}
\mathcal{L}_{\mathrm{bnd}}
=
-\frac{1}{N}
\sum_{i=1}^{N}
\left[
b_i\log\hat{b}_i
+
(1-b_i)\log(1-\hat{b}_i)
\right],
\label{eq:seg_bnd}
\end{equation}

\begin{equation}
\mathcal{L}_{\mathrm{seg}}
=
\mathcal{L}_{\mathrm{main}}
+
\lambda_{\mathrm{aux}}\mathcal{L}_{\mathrm{aux}}
+
\lambda_{\mathrm{bnd}}\mathcal{L}_{\mathrm{bnd}}.
\label{eq:seg_pretrain}
\end{equation}

Here, $P$ and $Y$ denote the main segmentation prediction and ground-truth target, respectively.
$P_s^{\mathrm{aux}}$ denotes the prediction from the $s$-th auxiliary head with target $Y_s$, and $S$ is the number of auxiliary heads.
$b_i$ and $\hat{b}_i$ are the ground-truth and predicted boundary values at pixel $i$, with $N$ denoting the number of pixels.
$\mathcal{L}_{\mathrm{wce}}$ and $\mathcal{L}_{\mathrm{dice}}$ denote the weighted cross-entropy and foreground Dice losses, while $\lambda_{\mathrm{aux}}$ and $\lambda_{\mathrm{bnd}}$ weight the corresponding auxiliary and boundary terms.

The resulting segmentation model is used to generate anatomical pseudo masks for Stage 2 and to initialize the Student network in Stage 3.

\subsection{Motion-Estimation Pretraining}
\subsubsection{Diffusion-Conditioned Motion Estimation Network}

Given $I_m$ and $I_f$, a perturbed fixed image $I_f^t$ is generated using the standard forward diffusion process~\citep{Ho2020}. For each training pair, a diffusion timestep $t$ and Gaussian noise $\epsilon \sim \mathcal{N}(0,\mathbf{I})$ are randomly sampled:
\begin{equation}
I_f^t
=
\sqrt{\bar{\alpha}_t}\, I_f
+
\sqrt{1-\bar{\alpha}_t}\, \epsilon,
\label{eq:forward_diffusion}
\end{equation}
where $\bar{\alpha}_t$ denotes the cumulative product of the noise-schedule coefficients. Following FSDiffReg~\citep{Qin2023}, the moving image, clean fixed image, and perturbed fixed image are concatenated as
\begin{equation}
X_t
=
\operatorname{Concat}\left(I_m, I_f, I_f^t\right).
\label{eq:diffusion_input}
\end{equation}

\begin{figure*}
    \centering
    \includegraphics[width=\textwidth]{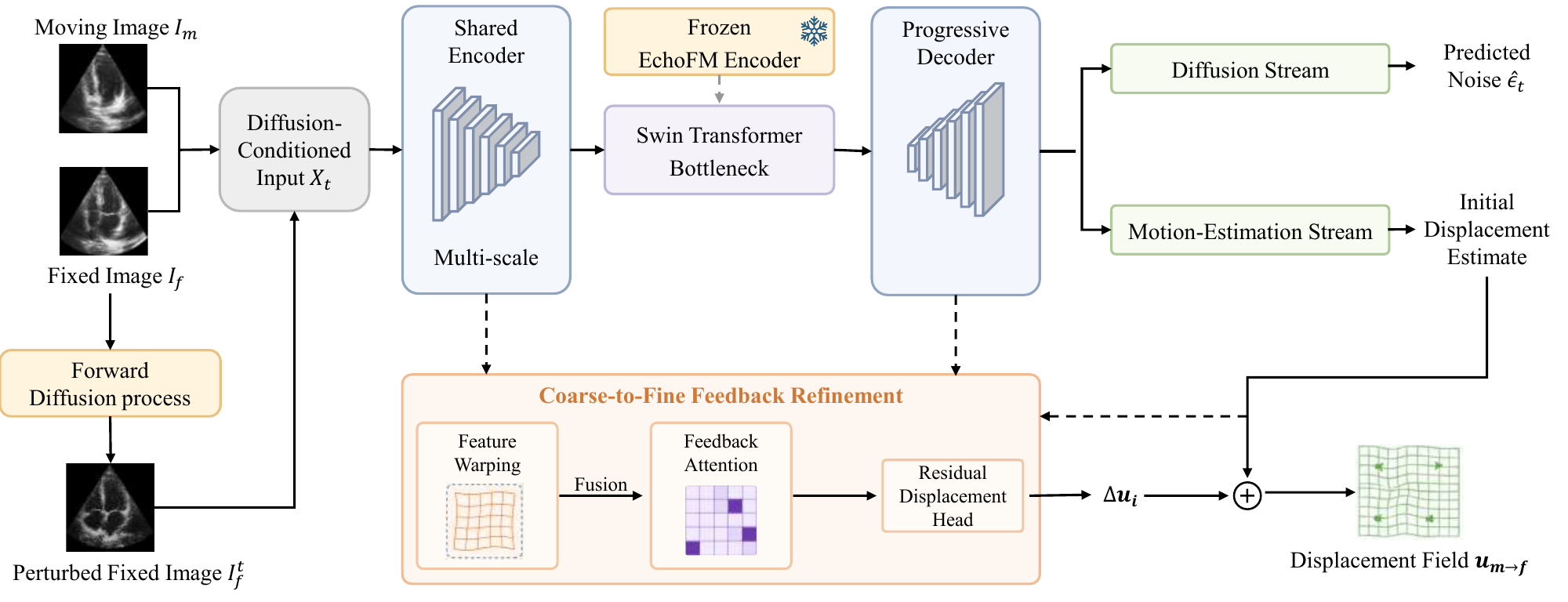}
    \caption{Architecture of the proposed diffusion-conditioned motion-estimation network. The moving image $I_m$, fixed image $I_f$, and perturbed fixed image $I_f^t$ form the diffusion-conditioned input $X_t$. The network contains a diffusion stream for noise prediction and a motion-estimation stream for motion estimation, together with coarse-to-fine feedback refinement. The diffusion timestep embedding is injected into the residual encoding blocks, and frozen EchoFM features are fused with the bottleneck feature through cross-attention.}
    \label{fig:motion_estimation}
\end{figure*}

The overall architecture of the proposed diffusion-condi\-tioned motion-estimation network is illustrated in Fig.~\ref{fig:motion_estimation}.

The sampled timestep $t$ is embedded and injected into the residual encoding blocks. A shared encoder extracts multi-scale diffusion-conditioned features from $X_t$, and the deepest feature map is further processed by a Swin Transformer V2 bottleneck~\citep{Liu2022} to capture long-range spatial dependencies.

To complement these learned features with echocardiography\allowbreak-specific structural information, the clean fixed image $I_f$ is additionally processed by a pretrained EchoFM encoder~\citep{Kim2025}.
The EchoFM encoder remains frozen during training and inference. Its high-level feature $F_E$ is projected to match the dimensionality of the bottleneck feature $F_B$ and fused with $F_B$ through cross-attention:
\begin{equation}
\widetilde{F}_B
=
\mathcal{F}_{\mathrm{CA}}(F_B,F_E),
\label{eq:cross_attention_fusion}
\end{equation}
where $\mathcal{F}_{\mathrm{CA}}(\cdot,\cdot)$ denotes the cross-attention fusion operation.

The fused bottleneck feature $\widetilde{F}_B$ is subsequently decoded through two task-specific streams. The diffusion stream predicts the injected noise $\hat{\epsilon}_t$, while the motion-estimation stream combines multi-scale decoder features with encoder skip connections to estimate the dense moving-to-fixed displacement field $\mathbf{u}_{m\to f}$. The displacement field is progressively refined in a coarse-to-fine manner and is finally used to warp the moving image toward the fixed image:
\begin{equation}
\hat{I}_f
=
\mathcal{W}
\left(
I_m,\mathbf{u}_{m\to f}
\right),
\label{eq:spatial_warping}
\end{equation}
where $\mathcal{W}(\cdot,\cdot)$ denotes differentiable spatial warping.

The stochastic diffusion perturbation is used only during training; inference does not require iterative reverse-diffusion sampling. During inference, $t$ is fixed to $0$ and the perturbed input is replaced by the clean fixed image, yielding $X_0=\operatorname{Concat}(I_m,I_f,I_f)$. The displacement field is therefore predicted through a single deterministic forward pass.

\subsubsection{Feedback Refinement}

Inspired by the feedback-attention strategy of~\citep{Hasan2025}, which uses residual registration errors after an initial alignment to guide subsequent refinement, we incorporate misalignment-aware feedback into the coarse-to-fine decoder. At decoder scale $i$, the displacement $\mathbf{u}_{i-1}$ and feedback $\overline{\mathcal{M}}_{i-1}$ propagated from the preceding coarser scale are resampled to the current resolution and combined with the current diffusion-stream decoder feature $\mathcal{F}_{i}^{\mathrm{diff}}$ to obtain a preliminary displacement:
\begin{equation}
\small
\mathbf{u}_{i}^{\mathrm{pre}}
=
\mathcal{U}\left(\mathbf{u}_{i-1}\right)
+
\mathcal{H}_{i}^{\mathrm{pre}}
\left(
\mathcal{F}_{i}^{\mathrm{diff}},
\mathcal{U}\left(\mathbf{u}_{i-1}\right),
\mathcal{U}\left(\overline{\mathcal{M}}_{i-1}\right)
\right),
\label{eq:preliminary_displacement}
\end{equation}
where $\mathcal{U}(\cdot)$, $\overline{(\cdot)}$, and $\mathcal{H}_{i}^{\mathrm{pre}}(\cdot)$ denote resampling, averaging, and the initial residual-flow head, respectively.
$\overline{\mathcal{M}}_{i-1}$ represents the average response of residual misalignment at the corresponding scale.

The preliminary displacement is then used to warp the corresponding encoder features. The original and warped features are integrated and subsequently processed by the feedback-attention
module to generate the feedback map $\mathcal{M}_{i,j}$:
\begin{equation}
\mathcal{M}_{i,j}
=
1-\mathcal{A}_{i,j}.
\label{eq:feedback_map}
\end{equation}

The scale-level feedback is then combined with the motion-estimation-stream decoder feature $\mathcal{F}_{i}^{\mathrm{mot}}$ and the preliminary displacement to refine the displacement:
\begin{equation}
\mathbf{u}_{i}
=
\mathbf{u}_{i}^{\mathrm{pre}}
+
\mathcal{H}_{i}^{\mathrm{ref}}
\left(
\mathcal{F}_{i}^{\mathrm{mot}},
\mathbf{u}_{i}^{\mathrm{pre}},
\overline{\mathcal{M}}_{i}
\right),
\label{eq:refined_displacement}
\end{equation}
where $\mathcal{H}_{i}^{\mathrm{ref}}(\cdot)$ denotes the residual-flow refinement head. The refined displacement $\mathbf{u}_{i}$ is propagated to the next finer scale, while a detached copy of $\overline{\mathcal{M}}_{i}$ serves as the feedback signal.

\subsubsection{Motion-Estimation Objective}

The motion-estimation objective combines diffusion noise prediction, image similarity, displacement-field regularization, and anatomical correspondence. The anatomical correspondence term is defined as the Dice loss between the warped moving pseudo mask and the corresponding fixed pseudo mask, both obtained from the pretrained segmentation network. The diffusion stream is supervised by the mean squared error between the sampled Gaussian noise $\epsilon$ and the predicted noise $\hat{\epsilon}_t$:
\begin{equation}
\mathcal{L}_{\mathrm{noise}}
=
\operatorname{MSE}
\left(
\epsilon,\hat{\epsilon}_t
\right).
\label{eq:noise_loss}
\end{equation}

The predicted noise is further used to modulate the image-similarity objective, following the diffusion-guided weighting strategy of FSDiffReg~\citep{Qin2023}. Specifically, it is transformed into a bounded spatial weighting map:
\begin{equation}
\omega_t
=
\left[
\sigma\left(\hat{\epsilon}_t\right)
\right]^{\gamma},
\label{eq:noise_weight}
\end{equation}
where $\sigma(\cdot)$ denotes the sigmoid function and $\gamma$ controls the weighting strength. Rather than treating $\hat{\epsilon}_t$ as an explicit score function, $\omega_t$ is used as a noise-prediction-guided weighting signal for local image similarity. Given the warped moving image
$\hat{I}_f=\mathcal{W}(I_m,\mathbf{u}_{m\to f})$, the weighted similarity loss is defined as
\begin{equation}
\mathcal{L}_{\mathrm{sim}}
=
-\operatorname{mean}
\left[
\omega_t
\odot
\mathrm{LCC}_{9\times9}
\left(
\hat{I}_f,I_f
\right)
\right],
\label{eq:similarity_loss}
\end{equation}
where $\mathrm{LCC}_{9\times9}$ denotes the local cross-correlation computed within a $9\times9$ window. Together with the displacement-field regularization and anatomical correspondence terms, \(\mathcal{L}_{\mathrm{noise}}\) and \(\mathcal{L}_{\mathrm{sim}}\) constitute the motion-estimation objective \(\mathcal{L}_{\mathrm{mot}}\).

\subsection{Self-distillation-based Joint Learning}

\begin{figure*}
    \centering
    \includegraphics[width=\textwidth]{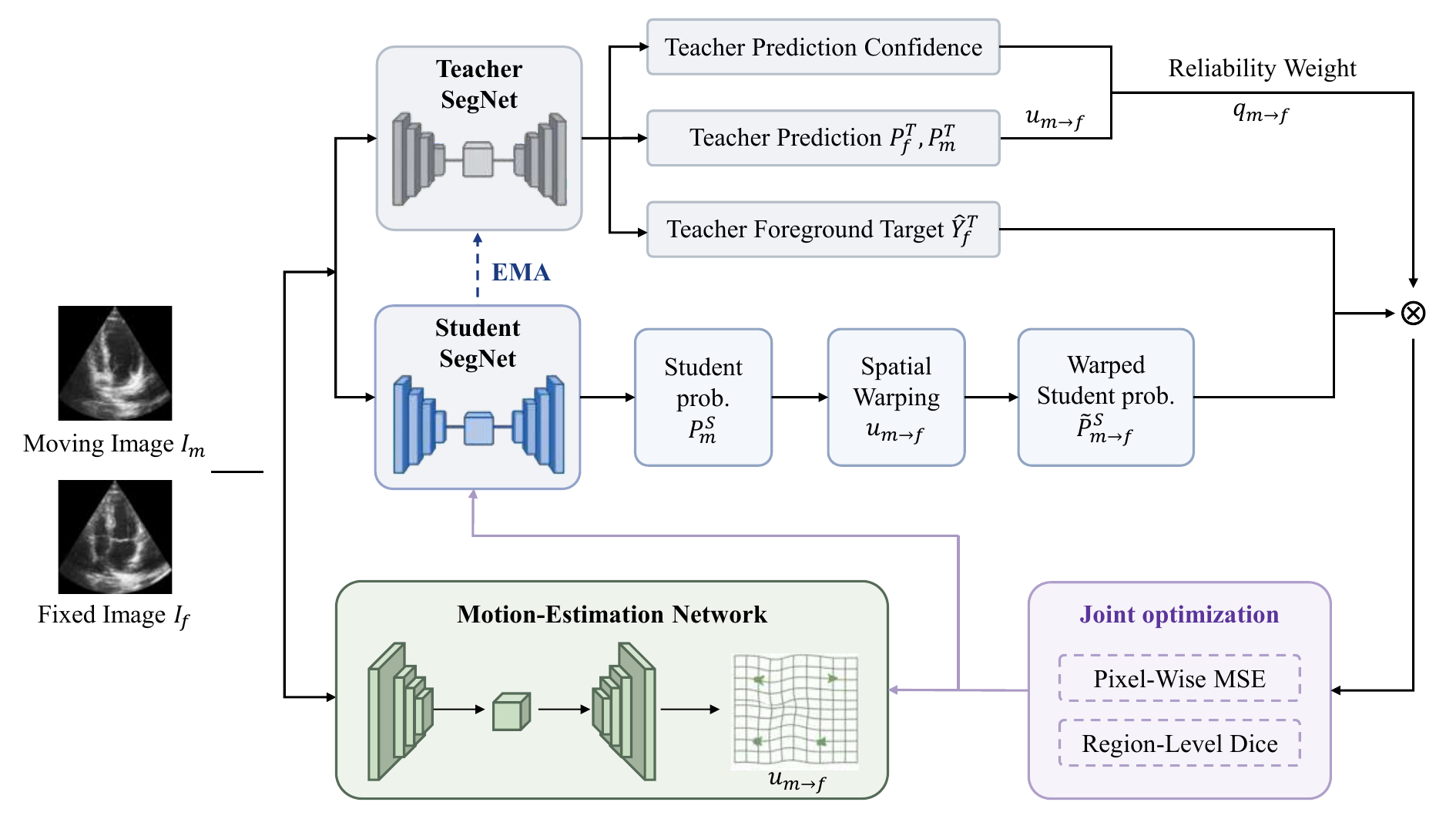}
    \caption{The proposed self-distillation-based joint learning strategy. Student and Teacher segmentation networks are employed, with the Teacher updated using EMA. Reliability-aware self-distillation and bidirectional anatomy–motion consistency are introduced in EchoDiST. Anatomical consistency is imposed in both moving-to-fixed and fixed-to-moving directions, while only the moving-to-fixed branch is shown for clarity. The Student prediction is warped to the fixed frame using the estimated displacement field and constrained by an EMA Teacher-derived anatomical target. A sample-level reliability weight modulates the combined region-level Dice and pixel-wise MSE consistency losses, enabling joint optimization of anatomical segmentations  and motion estimation.}
    \label{fig:joint_learning}
\end{figure*}

As illustrated in Fig.~\ref{fig:joint_learning}, anatomical segmentation and unsupervised motion estimation are jointly refined through EMA-based self-distillation and reliability-weighted cross-frame consistency.

To provide more stable anatomical guidance, we adopt an EMA-based self-distillation scheme following the Mean Teacher paradigm~\citep{Tarvainen2017}. Let $f_S(\cdot;\theta_S)$ and $f_T(\cdot;\theta_T)$ denote the Student and Teacher segmentation networks, respectively. The Student is initialized from the segmentation model pretrained in Stage 1, with the Teacher initialized using the same parameters. During joint training, the Student is updated by gradient descent, whereas the Teacher parameters are updated as an exponential moving average of the Student:
\begin{equation}
\theta_T
\leftarrow
\alpha \theta_T
+
(1-\alpha)\theta_S,
\label{eq:ema_update}
\end{equation}
where $\alpha$ denotes the EMA decay factor. The Teacher therefore provides EMA-stabilized anatomical targets without direct back-propagation.

Given $I_m$ and $I_f$, $f_S(\cdot;\theta_S)$ predicts the moving-frame foreground probability map $P_m^S$, while the motion-estimation network predicts the forward displacement field $\mathbf{u}_{m\to f}$. The probability map $P_m^S$ is then propagated to the fixed frame as
\begin{equation}
\widetilde{P}_{m\to f}^{S}
=
\mathcal{W}
\left(
P_m^S,
\mathbf{u}_{m\to f}
\right).
\label{eq:forward_probability_warp}
\end{equation}

Meanwhile, the Teacher prediction for $I_f$ is converted into a hard foreground target $\hat{Y}_f^T$.

The forward anatomical consistency loss is defined as
\begin{equation}
\mathcal{L}_{\mathrm{con}}^{m\to f}
=
\mathcal{L}_{\mathrm{dice}}
\left(
\widetilde{P}_{m\to f}^{S},
\hat{Y}_f^T
\right)
+
\lambda_c
\mathcal{L}_{\mathrm{mse}}
\left(
\widetilde{P}_{m\to f}^{S},
\hat{Y}_f^T
\right).
\label{eq:forward_anatomical_consistency}
\end{equation}

The Dice term constrains region-level anatomical overlap, while the MSE term further regularizes pixel-wise agreement. Because $\mathbf{u}_{m\to f}$ remains attached to the computational graph during warping, this consistency loss jointly optimizes anatomical segmentation and motion estimation. The same constraint is applied in the reverse direction to obtain $\mathcal{L}_{\mathrm{con}}^{f\to m}$.

Since Teacher-derived anatomical targets may vary in reliability across image pairs and motion directions, we further introduce a reliability-aware weighting strategy inspired by uncertainty-aware self-ensembling~\citep{Yu2019}. For each direction, three normalized reliability scores are computed from the Teacher outputs: $q_{\mathrm{conf}}$ measures the foreground-conditioned confidence of
the moving and fixed predictions, while $q_{\mathrm{agr}}$ and $q_{\mathrm{size}}$ respectively quantify the anatomical agreement and foreground-area consistency between the warped moving and fixed predictions.
The sample-level reliability is defined as
\begin{equation}
\small
q_{m\to f}
=
\frac{
\alpha_{\mathrm{conf}}q_{\mathrm{conf}}
+
\alpha_{\mathrm{agr}}q_{\mathrm{agr}}
+
\alpha_{\mathrm{size}}q_{\mathrm{size}}
}{
\alpha_{\mathrm{conf}}
+
\alpha_{\mathrm{agr}}
+
\alpha_{\mathrm{size}}
},
\
q_{m\to f}\in[0,1].
\label{eq:reliability_forward}
\end{equation}

The same formulation is applied to the reverse direction to obtain $q_{f\to m}$. Each component is mapped to $[0,1]$ before weighted aggregation. The displacement field used for reliability estimation is detached from the computational graph, and the resulting reliability score is treated as a stop-gradient sample weight in the subsequent anatomical consistency objective.

The overall joint-training objective is
\begin{equation}
\small
\mathcal{L}_{\mathrm{joint}}
=
\mathcal{L}_{\mathrm{mot}}
+
\lambda_{\mathrm{seg}}\mathcal{L}_{\mathrm{seg}}
+
q_{m\to f}\lambda_{\mathrm{mf}}
\mathcal{L}_{\mathrm{con}}^{m\to f}
+
q_{f\to m}\lambda_{\mathrm{fm}}
\mathcal{L}_{\mathrm{con}}^{f\to m}.
\label{eq:joint_objective}
\end{equation}

\section{Experiments}
\subsection{Datasets and Preprocessing}

Three echocardiographic datasets were used in this study, comprising two clinical cohorts, i.e., CAMUS~\citep{Leclerc2019} and EchoNet-Dynamic~\citep{Ouyang2020}, and a CAMUS simulation dataset~\citep{Evain2022}. The CAMUS dataset was used for model development, with its apical four-chamber (4CH) sequences used for training and in-domain evaluation and its apical two-chamber (2CH) sequences used for cross-view evaluation. The CAMUS simulation dataset and EchoNet-Dynamic dataset were treated as external test datasets and used to evaluate cross-dataset generalization without dataset-specific fine-tuning. All images in the three datasets were resized to $256\times256$ pixels using bilinear interpolation. Grayscale intensities were scaled to $[0,1]$ for anatomical segmentation. For motion-estimation training, each image was independently intensity-normalized and linearly mapped to $[-1,1]$. Unless otherwise specified, motion estimation was performed from end-diastole (ED) to end-systole (ES), with the ED frame treated as the moving frame and the ES frame as the fixed frame. For full-cycle analyses, ED served as the reference frame, and motion was estimated from ED to each frame throughout the cardiac cycle.

\textbf{CAMUS.} The CAMUS dataset~\citep{Leclerc2019} comprises apical two-chamber (2CH) and four-chamber (4CH) echocardiographic sequences from 500 patients, with manual delineations of the left ventricular (LV) cavity, myocardium, and left atrium (LA) provided at ED and ES. The CAMUS-4CH cohort was split at the patient level into 350 training, 50 validation, and 100 held-out test subjects. Of the 350 training subjects, 50 with manual anatomical annotations were used for segmentation pretraining, while all 350 subjects were used for motion-estimation training. The corresponding 100-subject CAMUS-2CH test cohort was used exclusively for cross-view evaluation without additional fine-tuning.

\textbf{EchoNet-Dynamic.} EchoNet-Dynamic~\citep{Ouyang2020} comprises 10,030 apical 4CH echocardiographic videos with cardiac functional measurements and expert LV tracings at ED and ES. We used a fixed subset of 110 cases with complete videos and corresponding LV tracings. This subset was defined independently of model predictions and downstream evaluation metrics. The official LV tracings were converted into closed contours and binary cavity masks for evaluation. The annotated frames with the largest and smallest LV cavity areas were identified as ED and ES, respectively.

\begin{figure}
    \centering
    \includegraphics[width=\columnwidth]{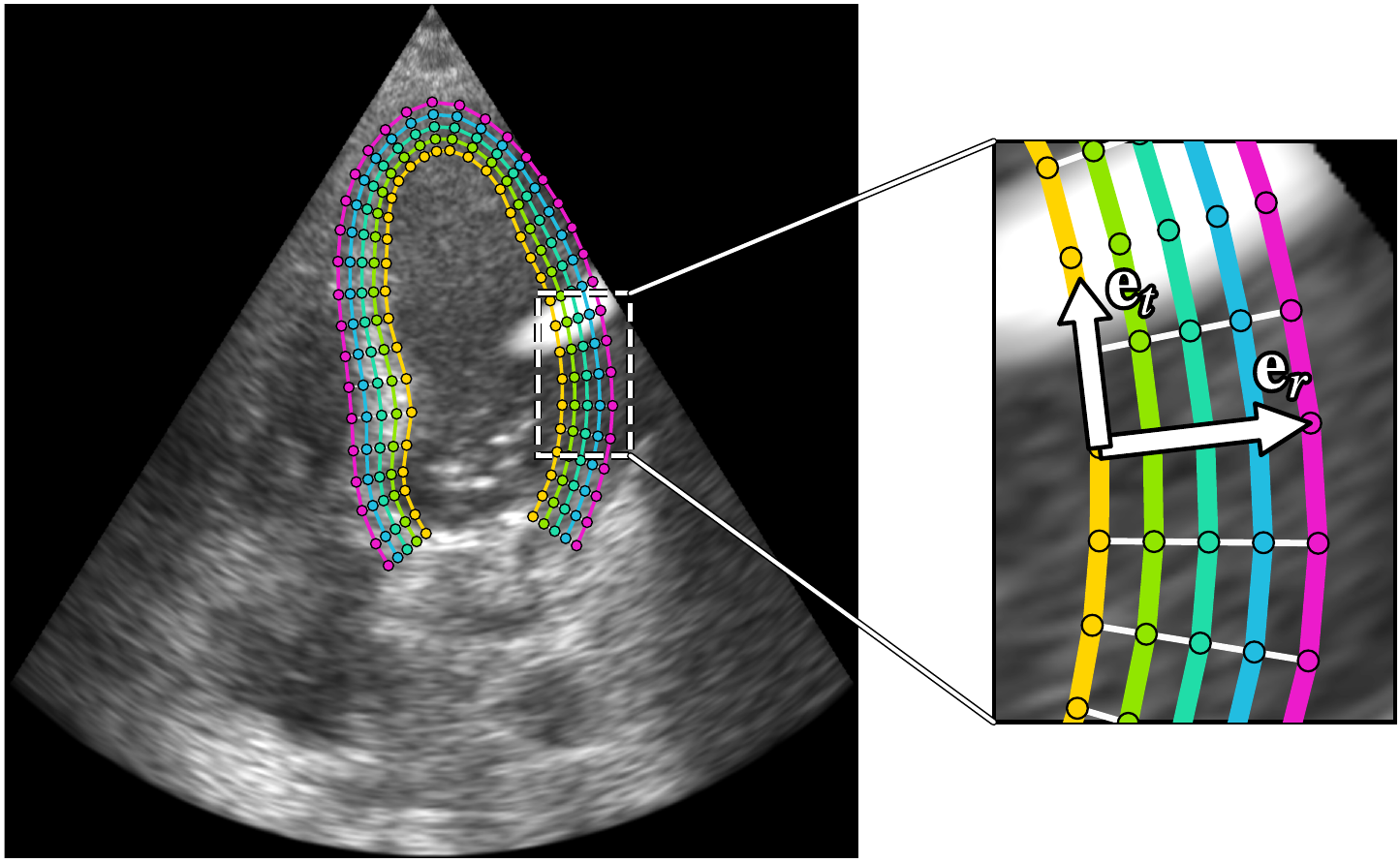}
    \caption{Reference myocardial point configuration in the CAMUS simulation dataset, comprising five transmural layers with 36 points distributed along each layer. The enlarged region illustrates the local radial-tangential basis, where the radial direction $\mathbf{e}_r$ is defined from the corresponding inner-layer point toward the outer-layer point, and the tangential direction $\mathbf{e}_t$ is defined as the in-plane direction orthogonal to $\mathbf{e}_r$.}
    \label{fig:simulation}
\end{figure}

\textbf{CAMUS simulation dataset.} The CAMUS simulation dataset~\citep{Evain2022} contains simulated apical 4CH echocardiographic sequences with reverberation artifacts for 100 patients from the CAMUS dataset. Each case comprises a B-mode image sequence, time-varying myocardial meshes, and ED/ES frame indices, from which reference myocardial motion throughout the cardiac cycle can be derived. The myocardial meshes are represented by 180 points organized into five transmural layers at each frame, with 36 points distributed circumferentially along each layer. We denote each point as $p_{i,j}$, where $i \in \{0,\ldots,4\}$ indexes the transmural layer from the inner to the outer myocardial boundary and $j \in \{0,\ldots,35\}$ denotes the circumferential position within each layer. Points sharing the same circumferential index $j$ across different layers define corresponding transmural locations, as illustrated in Fig.~\ref{fig:simulation}.

\subsection{Implementation Details}

The experiments were conducted on an RTX A6000-equipped server for training and evaluation, with all network code implemented in PyTorch. 

Training followed the three-stage strategy described in Section~\ref{sec:Method}. In Stage 1, the anatomical segmentation network was pretrained for 100 epochs using AdamW with a batch size of 8 and a weight decay of $1\times10^{-4}$. The learning rates were set to $1\times10^{-5}$ for the pretrained EchoFM backbone and $1\times10^{-4}$ for the segmentation heads. The motion-estimation network was optimized for 100 epochs in both Stage 2 and Stage 3 using Adam with a learning rate of $1\times10^{-4}$ and a batch size of 1. During Stage 2, the forward diffusion process used 1000 timesteps with a linear variance schedule from $1\times10^{-6}$ to $1\times10^{-2}$. In Stage 3, the Student segmentation network was optimized using AdamW with a learning rate of $1\times10^{-4}$, a weight decay of $1\times10^{-4}$, and a batch size of 8, while the Teacher network was updated after each optimization step using EMA with a decay factor of 0.995. 

The EchoFM encoder in the motion-estimation network remained frozen during Stages 2 and 3. At inference, motion estimation was performed by a single deterministic forward pass without reverse-diffusion sampling.

\subsection{Evaluation Metrics}
\subsubsection{Anatomical Alignment}

In our experiments, we employed three metrics to assess anatomical alignment performance: the Dice similarity coefficient (Dice)~\citep{Dice1945}, the 95th-percentile Hausdorff distance (HD95)~\citep{Huttenlocher1993}, and the average symmetric surface distance (ASSD)~\citep{Heimann2009}. Metrics were averaged over the LV cavity, myocardium, and LA for CAMUS; for the CAMUS simulation dataset and EchoNet-Dynamic, they were computed on the myocardium and LV cavity, respectively. HD95 and ASSD were reported in millimeters for CAMUS and the CAMUS simulation dataset and in pixels for EchoNet-Dynamic. A better alignment is characterized by a higher Dice score and lower HD95 and ASSD values.

For the CAMUS simulation dataset with available myocardial point correspondences, endpoint error (EPE) was used to quantify motion accuracy~\citep{Baker2011}. EPE was defined as the Euclidean distance between corresponding predicted and reference points, computed in physical coordinates and reported in millimeters.

\subsubsection{Myocardial Strain Assessment}

Myocardial strain was evaluated on the CAMUS simulation dataset at both local and global levels.

\textbf{Local myocardial strain.}
Local myocardial strain was quantified in the ED-defined radial-tangential coordinate system illustrated in Fig.~\ref{fig:simulation}. At each myocardial position, the radial direction was defined from the corresponding inner-layer point toward the outer-layer point, while the tangential direction was defined as the in-plane direction orthogonal to the radial direction. Consistent with previous echocardiographic motion studies that derive directional myocardial strain from estimated motion fields in a cardiac coordinate system~\citep{Ahn2023}, local displacement gradients were used to derive the Green-Lagrange strain tensor, which was then transformed from the axial-lateral image coordinates into the local radial-tangential coordinate system. The radial and tangential components were used for evaluation, while the associated shear component was not included in the reported metrics. The same procedure was applied to the predicted and reference motion fields. ES radial- and tangential-strain errors were quantified over the 180 myocardial points using root mean square error (RMSE) and reported in percent (\%).

\textbf{Global myocardial strain.} ED was used as the reference frame for full-cycle strain estimation, consistent with standard echocardiographic displacement analysis~\citep{MorAvi2011,Voigt2015}. Myocardial motion was estimated directly from ED to each subsequent frame. Global tangential strain was defined as the spatial average of local tangential strain over the myocardium at each frame. Agreement between predicted and reference strain trajectories was evaluated using curve RMSE and Pearson correlation $r$, with RMSE reported in percent (\%).

Global longitudinal strain (GLS) was derived geometrically from the relative change in endocardial contour length with respect to its ED reference length. The endocardial contour was represented as an open curve extending from one mitral annular point, through the ventricular apex, to the opposite mitral annular point, without closure across the mitral valve plane. Predicted and reference GLS trajectories were compared using curve RMSE and Pearson correlation $r$, with RMSE reported in percent (\%).

\subsubsection{Motion-Derived Functional and Cardiac-Phase Assessment}

\textbf{Area-based functional accuracy.} On EchoNet-Dynamic, we assessed whether the estimated motion preserved LV systolic contraction using the available expert ED/ES annotation. Similar to the relative ED-to-ES volume change used in conventional ejection fraction, an area-based ejection fraction (Area-EF) was defined as
\begin{equation}
\mathrm{Area\text{-}EF}
=
\frac{A_{\mathrm{ED}}-A_{\mathrm{ES}}}{A_{\mathrm{ED}}}
\times 100\%.
\label{eq:area_ef}
\end{equation}

Reference Area-EF was computed from the annotated ED and ES cavity areas, whereas predicted Area-EF was computed from the ED cavity and its motion-propagated ES area. Agreement was evaluated using RMSE and Pearson correlation $r$, with RMSE reported in percent (\%).

\textbf{ED/ES frame prediction.} We further assessed whether the proposed method can identify the ED and ES frames. Given an annotated LV mask, the estimated displacement fields were used to propagate the mask to different time points and obtain the corresponding LV-area trajectory. Since the LV area is generally largest at ED and smallest at ES, the frames with the maximum and minimum predicted LV areas were identified as the predicted ED and ES frames, respectively. ES localization error quantified the absolute frame difference between the annotated ES frame and the frame of minimum predicted LV area within the ED-to-ES interval, while monotonic consistency measured the proportion of consecutive frame transitions over this interval for which the predicted LV area did not increase.

\subsubsection{Statistical Analysis}
Statistical comparisons were conducted at the case level. Continuous paired endpoints were analyzed using two-sided Wilcoxon signed-rank tests. Multiple comparisons were corrected using the Holm procedure, and adjusted $p$-values below $0.05$ were considered statistically significant. Differences between dependent Pearson correlations on EchoNet-Dynamic were assessed using paired case-level bootstrap resampling.

\subsection{Comparative Evaluation}
\subsubsection{Comparison Methods}

We compared EchoDiST with seven representative learning-based methods: DdC-AC-DLIR~\citep{Hasan2025}, NICE-Trans~\citep{Meng2023}, NICE-Net~\citep{Meng2022}, CorrMLP~\citep{Meng2024}, SDHNet~\citep{Zhou2023}, TransMorph2D~\citep{Chen2022}, and VoxelMorph~\citep{Balakrishnan2019}. For a fair comparison, methods requiring retraining were trained using the same CAMUS-4CH training dataset as EchoDiST, while their original method-specific objectives and optimization settings were retained.

\subsubsection{Anatomical Alignment and Generalization}

\begin{table*}[!t]
\centering
\caption{Quantitative comparison of anatomical alignment on CAMUS-4CH, CAMUS-2CH, EchoNet-Dynamic, and the CAMUS simulation dataset. All metrics are reported as mean $\pm$ standard deviation over 100, 100, 110, and 100 test cases for the four datasets, respectively, with one ED-to-ES image pair evaluated per case. $p$-values indicate paired statistical comparisons of Dice between EchoDiST and each comparison method.}
\label{tab:comparison_dice}

{\fontfamily{ptm}\selectfont
\small
\setlength{\tabcolsep}{4.2pt}

\begin{tabular}{lcccccccc}
\toprule
& \multicolumn{4}{c}{CAMUS-4CH}
& \multicolumn{4}{c}{CAMUS-2CH} \\
\cmidrule(lr){2-5}
\cmidrule(lr){6-9}

Method
& Dice$\uparrow$
& HD95$\downarrow$
& ASSD$\downarrow$
& $p$-value
& Dice$\uparrow$
& HD95$\downarrow$
& ASSD$\downarrow$
& $p$-value \\
\midrule

DdC-AC-DLIR
& \textit{0.886 $\pm$ 0.033}
& \textit{7.47 $\pm$ 2.72}
& \textit{2.88 $\pm$ 0.95}
& $<0.001$
& \textit{0.872 $\pm$ 0.039}
& \textit{8.66 $\pm$ 2.98}
& \textit{3.44 $\pm$ 1.16}
& $<0.001$ \\

NICE-Trans
& 0.863 $\pm$ 0.049
& 9.25 $\pm$ 3.89
& 3.46 $\pm$ 1.55
& $<0.001$
& 0.848 $\pm$ 0.052
& 10.43 $\pm$ 4.18
& 4.07 $\pm$ 1.65
& $<0.001$ \\

NICE-Net
& 0.849 $\pm$ 0.048
& 9.92 $\pm$ 3.85
& 3.70 $\pm$ 1.43
& $<0.001$
& 0.839 $\pm$ 0.052
& 11.05 $\pm$ 4.28
& 4.22 $\pm$ 1.56
& $<0.001$ \\

CorrMLP
& 0.848 $\pm$ 0.053
& 10.22 $\pm$ 4.17
& 3.83 $\pm$ 1.65
& $<0.001$
& 0.835 $\pm$ 0.057
& 10.96 $\pm$ 4.08
& 4.37 $\pm$ 1.76
& $<0.001$ \\

SDHNet
& 0.868 $\pm$ 0.047
& 9.73 $\pm$ 4.40
& 3.42 $\pm$ 1.28
& $<0.001$
& 0.861 $\pm$ 0.046
& 10.23 $\pm$ 4.27
& 3.79 $\pm$ 1.38
& $<0.001$ \\

TransMorph2D
& 0.849 $\pm$ 0.055
& 10.73 $\pm$ 4.80
& 4.03 $\pm$ 1.69
& $<0.001$
& 0.845 $\pm$ 0.054
& 10.92 $\pm$ 4.51
& 4.32 $\pm$ 1.69
& $<0.001$ \\

VoxelMorph
& 0.824 $\pm$ 0.051
& 12.75 $\pm$ 4.39
& 4.64 $\pm$ 1.57
& $<0.001$
& 0.819 $\pm$ 0.052
& 12.87 $\pm$ 4.41
& 5.02 $\pm$ 1.65
& $<0.001$ \\


\textbf{Ours}
& \textbf{0.909 $\pm$ 0.022}
& \textbf{5.96 $\pm$ 1.80}
& \textbf{2.29 $\pm$ 0.68}
& ref.
& \textbf{0.892 $\pm$ 0.032}
& \textbf{7.47 $\pm$ 2.65}
& \textbf{2.89 $\pm$ 1.03}
& ref. \\

\midrule

& \multicolumn{4}{c}{EchoNet-Dynamic}
& \multicolumn{4}{c}{CAMUS Simulation} \\
\cmidrule(lr){2-5}
\cmidrule(lr){6-9}

Method
& Dice$\uparrow$
& HD95$\downarrow$
& ASSD$\downarrow$
& $p$-value
& Dice$\uparrow$
& HD95$\downarrow$
& ASSD$\downarrow$
& $p$-value \\
\midrule

DdC-AC-DLIR
& 0.781 $\pm$ 0.077
& 7.37 $\pm$ 2.34
& 3.31 $\pm$ 1.13
& $<0.001$
& 0.781 $\pm$ 0.060
& 5.70 $\pm$ 1.69
& 2.12 $\pm$ 0.58
& $<0.001$ \\

NICE-Trans
& 0.783 $\pm$ 0.074
& 7.77 $\pm$ 2.50
& 3.21 $\pm$ 1.15
& $<0.001$
& \textit{0.820 $\pm$ 0.059}
& \textit{5.12 $\pm$ 1.62}
& \textit{1.73 $\pm$ 0.55}
& $<0.001$ \\

NICE-Net
& 0.760 $\pm$ 0.075
& 7.55 $\pm$ 2.37
& 3.65 $\pm$ 1.16
& $<0.001$
& 0.782 $\pm$ 0.057
& 5.71 $\pm$ 1.48
& 2.15 $\pm$ 0.54
& $<0.001$ \\

CorrMLP
& 0.784 $\pm$ 0.075
& 7.70 $\pm$ 2.48
& 3.14 $\pm$ 1.14
& $<0.001$
& 0.797 $\pm$ 0.058
& 5.80 $\pm$ 1.65
& 1.94 $\pm$ 0.56
& $<0.001$ \\

SDHNet
& 0.707 $\pm$ 0.086
& 9.14 $\pm$ 2.53
& 4.39 $\pm$ 1.28
& $<0.001$
& 0.689 $\pm$ 0.069
& 7.61 $\pm$ 1.83
& 2.97 $\pm$ 0.60
& $<0.001$ \\

TransMorph2D
& \textit{0.835 $\pm$ 0.071}
& \textit{6.36 $\pm$ 2.63}
& \textit{2.50 $\pm$ 1.08}
& $<0.001$
& 0.806 $\pm$ 0.054
& 5.61 $\pm$ 1.89
& 1.98 $\pm$ 0.57
& $<0.001$ \\

VoxelMorph
& 0.779 $\pm$ 0.072
& 7.39 $\pm$ 2.45
& 3.39 $\pm$ 1.15
& $<0.001$
& 0.772 $\pm$ 0.061
& 5.78 $\pm$ 1.58
& 2.23 $\pm$ 0.56
& $<0.001$ \\


\textbf{Ours}
& \textbf{0.888 $\pm$ 0.054}
& \textbf{4.19 $\pm$ 2.04}
& \textbf{1.67 $\pm$ 0.84}
& ref.
& \textbf{0.878 $\pm$ 0.037}
& \textbf{3.58 $\pm$ 1.24}
& \textbf{1.20 $\pm$ 0.33}
& ref. \\

\bottomrule
\end{tabular}
}
\end{table*}

The anatomical alignment performance of various methods across four datasets is detailed in Table~\ref{tab:comparison_dice}. For alignment accuracy, our EchoDiST achieves the highest Dice and the lowest HD95 and ASSD across all four evaluation settings. DdC-AC-DLIR achieves the strongest baseline performance on CAMUS, whereas NICE-Trans and TransMorph2D are more competitive on the CAMUS simulation dataset and EchoNet-Dynamic, respectively. In contrast, our approach consistently outperforms on all metrics, demonstrating more robust anatomical alignment across different views and datasets. Subject-level paired statistical analysis further confirms that the improvements over all comparison methods are significant for all three metrics on each dataset (all Holm-adjusted $p<0.001$).

\begin{figure*}[!t]
    \centering
    \includegraphics[width=\textwidth,trim=0 3.7cm 0 4.0cm,clip]{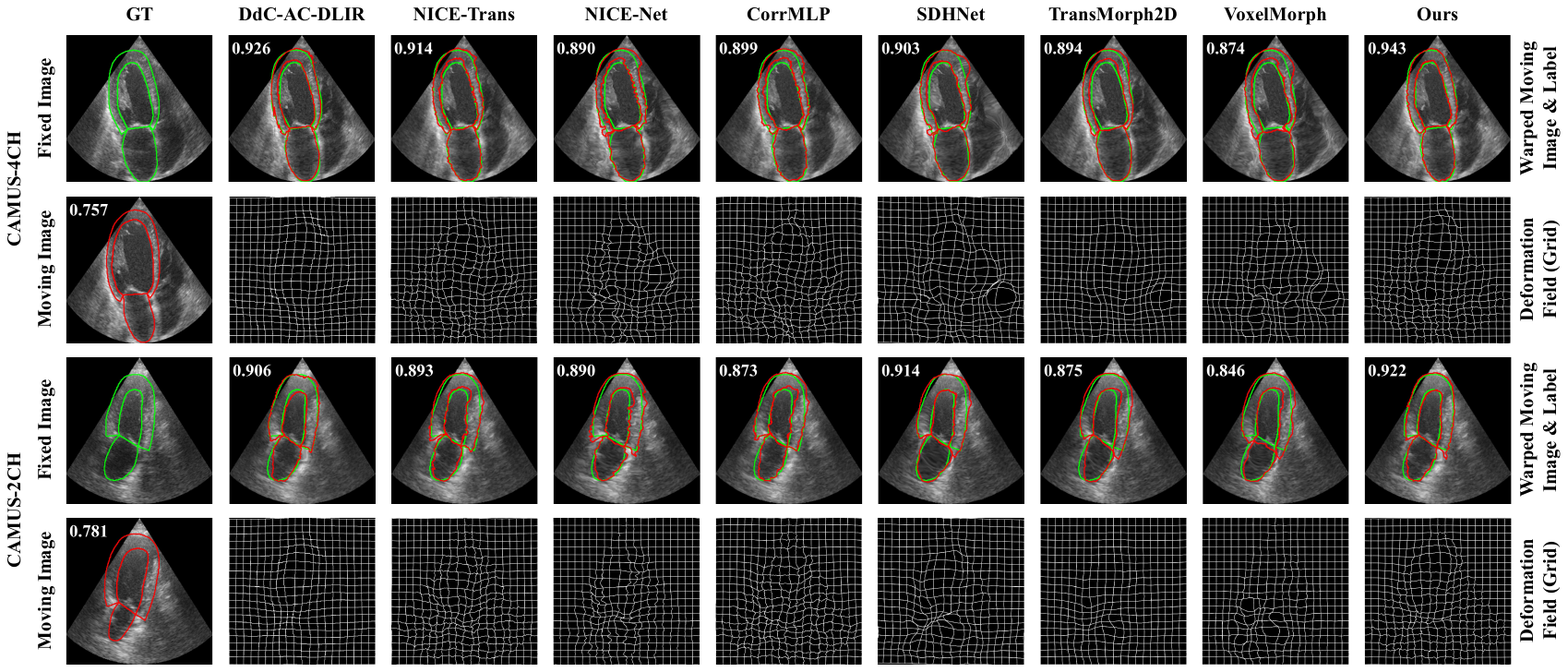}
    \caption{Qualitative comparison of different methods on the CAMUS dataset. The first column shows the ES fixed image and the ED moving image with anatomical contours, and the remaining columns show the results of the different methods. The dataset corresponding to each two-row group is indicated along the left margin. Within each group, the first row presents the warped moving image with the fixed anatomy in green and the warped moving anatomy in red, while the second row presents the corresponding displacement grid. The numbers displayed on the upper-left corner of each warped-image panel indicate the Dice score for the representative ED-to-ES image pair shown.}
    \label{fig:comparison_dice_1}
\end{figure*}

\begin{figure*}[!b]
    \centering
    \includegraphics[width=\textwidth,trim=0 3.7cm 0 4.0cm,clip]{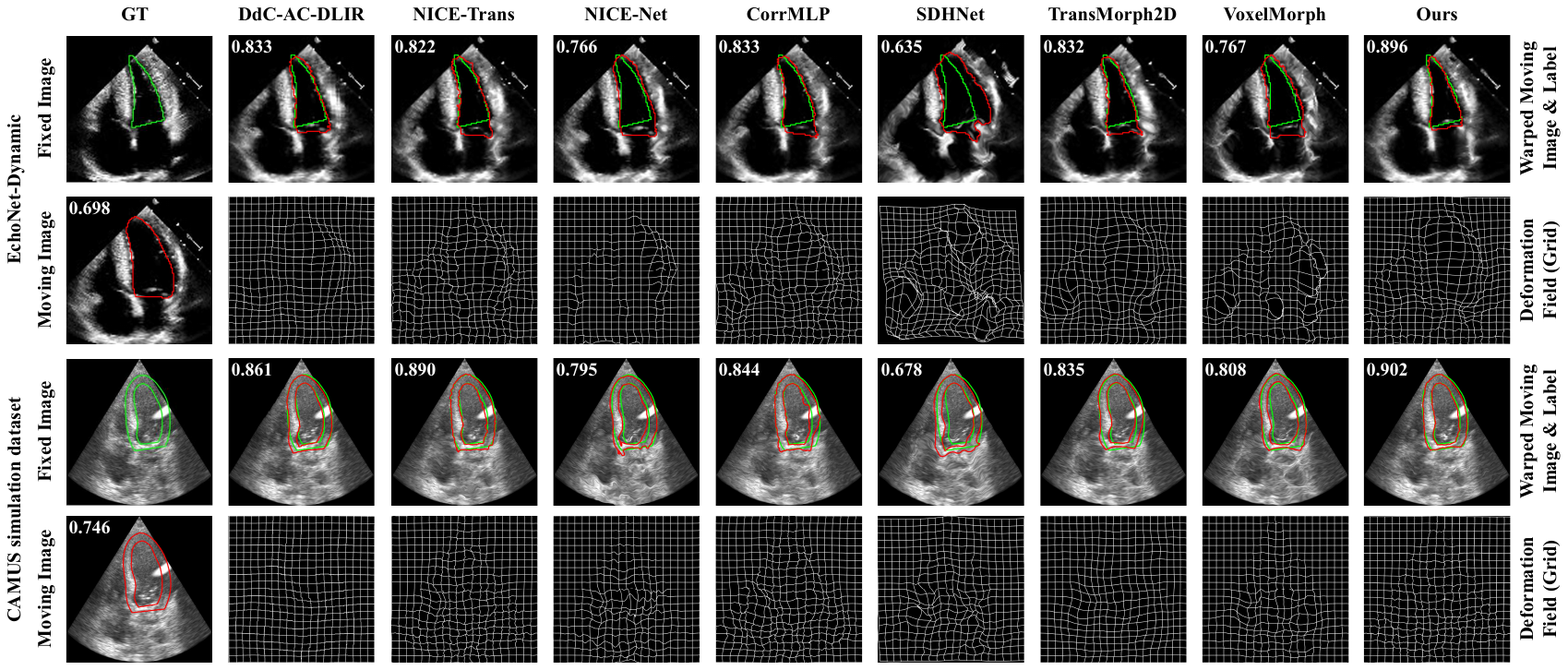}
    \caption{Qualitative comparison of different methods on EchoNet-Dynamic and the CAMUS simulation dataset. The first column shows the ES fixed image and the ED moving image with anatomical contours, and the remaining columns show the results of the different methods. The dataset corresponding to each two-row group is indicated along the left margin. Within each group, the first row presents the warped moving image with the fixed anatomy in green and the warped moving anatomy in red, while the second row presents the corresponding displacement grid. The numbers displayed on the upper-left corner of each warped-image panel indicate the Dice score for the representative ED-to-ES image pair shown.}
    \label{fig:comparison_dice_2}
\end{figure*}

Fig.~\ref{fig:comparison_dice_1} and~\ref{fig:comparison_dice_2} visually compare the alignment results of different methods, showing that EchoDiST achieves closer agreement between the warped and target anatomical boundaries, particularly around the ventricular wall and apex.

\subsubsection{Myocardial Motion and Strain Assessment}

\begin{table*}[!t]
\centering
\caption{Quantitative comparison of myocardial point motion and local strain estimation on the CAMUS simulation  dataset. All metrics are reported as mean $\pm$ standard deviation over 100 test cases, with 180 myocardial points evaluated per case. $p$-values indicate paired statistical comparisons of EPE between EchoDiST and each comparison method.}
\label{tab:comparison_local}

{\fontfamily{ptm}\selectfont
\small
\setlength{\tabcolsep}{7pt}

\begin{tabular}{lcccc}
\toprule
Method
& EPE (mm)$\downarrow$
& Radial RMSE (\%)$\downarrow$
& Tangential RMSE (\%)$\downarrow$
& $p$-value \\
\midrule

DdC-AC-DLIR
& 7.10 $\pm$ 2.31
& \textit{17.46 $\pm$ 5.52}
& 18.43 $\pm$ 5.83
& $<0.001$ \\

NICE-Trans
& \textit{6.33 $\pm$ 2.45}
& 24.56 $\pm$ 6.11
& 19.78 $\pm$ 5.65
& $<0.001$ \\

NICE-Net
& 7.31 $\pm$ 2.20
& 21.64 $\pm$ 5.49
& 22.61 $\pm$ 5.73
& $<0.001$ \\

CorrMLP
& 6.81 $\pm$ 2.51
& 27.77 $\pm$ 6.49
& 21.45 $\pm$ 5.59
& $<0.001$ \\

SDHNet
& 9.04 $\pm$ 2.23
& 25.61 $\pm$ 6.53
& 25.02 $\pm$ 6.32
& $<0.001$ \\

TransMorph2D
& 6.74 $\pm$ 2.42
& 20.40 $\pm$ 5.26
& \textit{18.67 $\pm$ 5.91}
& $<0.001$ \\

VoxelMorph
& 7.55 $\pm$ 2.30
& 25.04 $\pm$ 6.46
& 22.86 $\pm$ 6.17
& $<0.001$ \\

\textbf{Ours}
& \textbf{4.98 $\pm$ 1.58}
& \textbf{16.59 $\pm$ 4.27}
& \textbf{16.50 $\pm$ 5.33}
& ref. \\

\bottomrule
\end{tabular}
}
\end{table*}

Table~\ref{tab:comparison_local} reports the quantitative results of myocardial point motion and local strain estimation on the CAMUS simulation dataset. EchoDiST achieved the lowest point-motion error as well as the lowest radial- and tangential-strain errors, indicating consistently improved accuracy in both myocardial motion tracking and local strain estimation. Subject-level comparisons further showed significant improvements in EPE over all comparison methods ($p<0.001$), with significant improvements also observed for tangential strain ($p<0.05$).

\begin{figure*}
    \centering
    \includegraphics[width=0.9\textwidth,trim=0.2cm 2.2cm 0 2.3cm,clip]{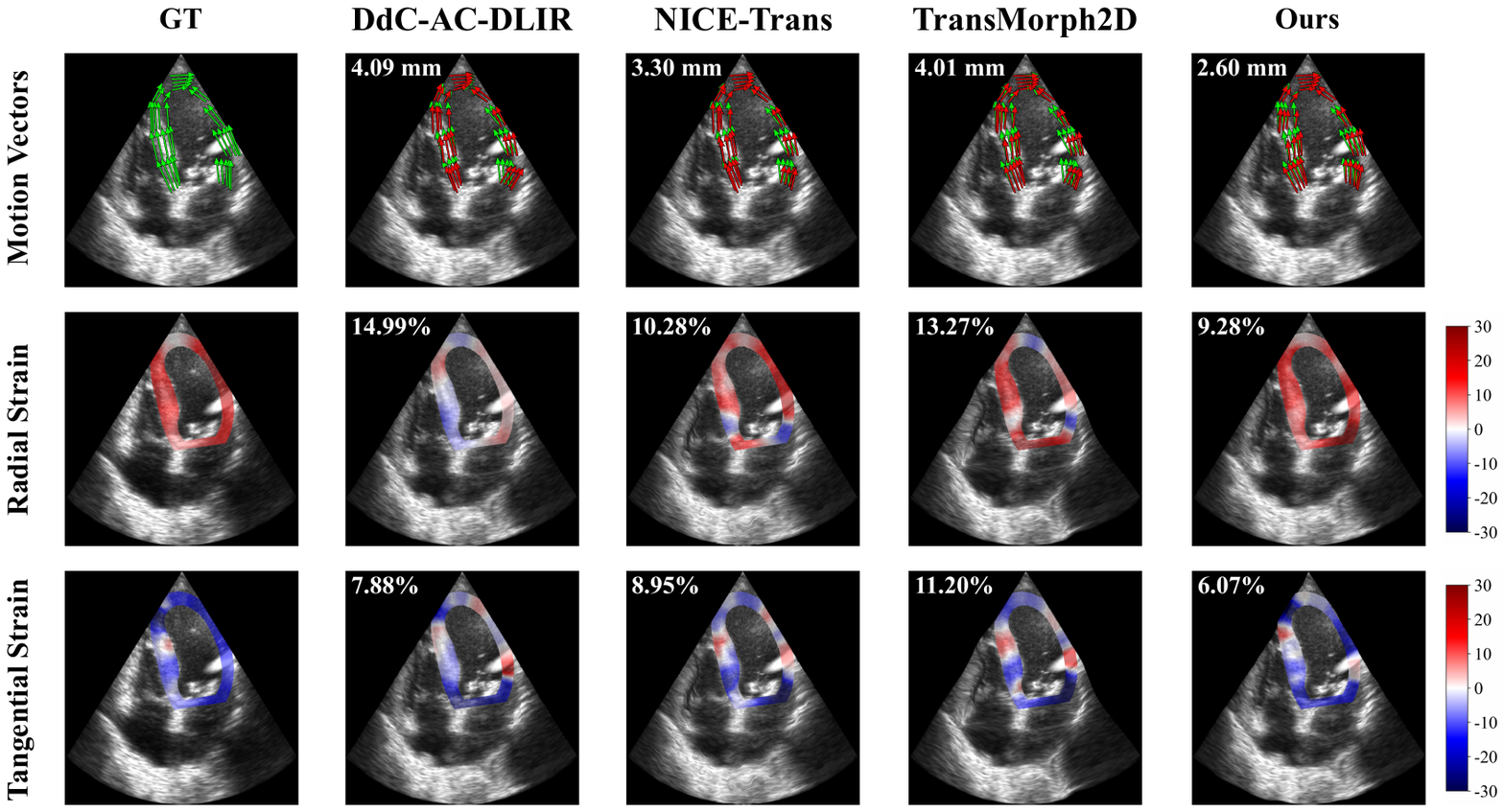}
    \caption{Qualitative comparison of selected representative methods for myocardial point motion and local strain estimation on the CAMUS simulation dataset. The first row shows ED-to-ES myocardial point-motion vectors, with reference and predicted motions shown in green and red, respectively. The second and third rows show the corresponding radial and tangential strain maps. EPE and strain RMSE are computed over the 180 myocardial points of the representative case and displayed on the upper-left corner of the corresponding motion and strain panels, respectively.}
    \label{fig:comparison_local}
\end{figure*}

The qualitative results obtained with selected representative methods in Fig.~\ref{fig:comparison_local} further support these findings. EchoDiST more closely reproduces the reference myocardial motion in both displacement direction and magnitude, while preserving the spatial organization of motion across the myocardial wall. The corresponding radial and tangential strain maps also show spatial patterns and regional variations that are more consistent with the reference distributions, whereas the comparison methods exhibit larger local discrepancies.

The advantage extends to full-cycle global strain assessment. As reported in Table~\ref{tab:comparison_global}, EchoDiST achieved the lowest curve RMSE and the highest Pearson correlation for both GTS and GLS, demonstrating closer agreement with the reference global strain trajectories. These improvements were statistically significant compared with all comparison methods for both global strain measures ($p<0.001$).

\begin{figure*}
    \centering
    \includegraphics[width=\textwidth]{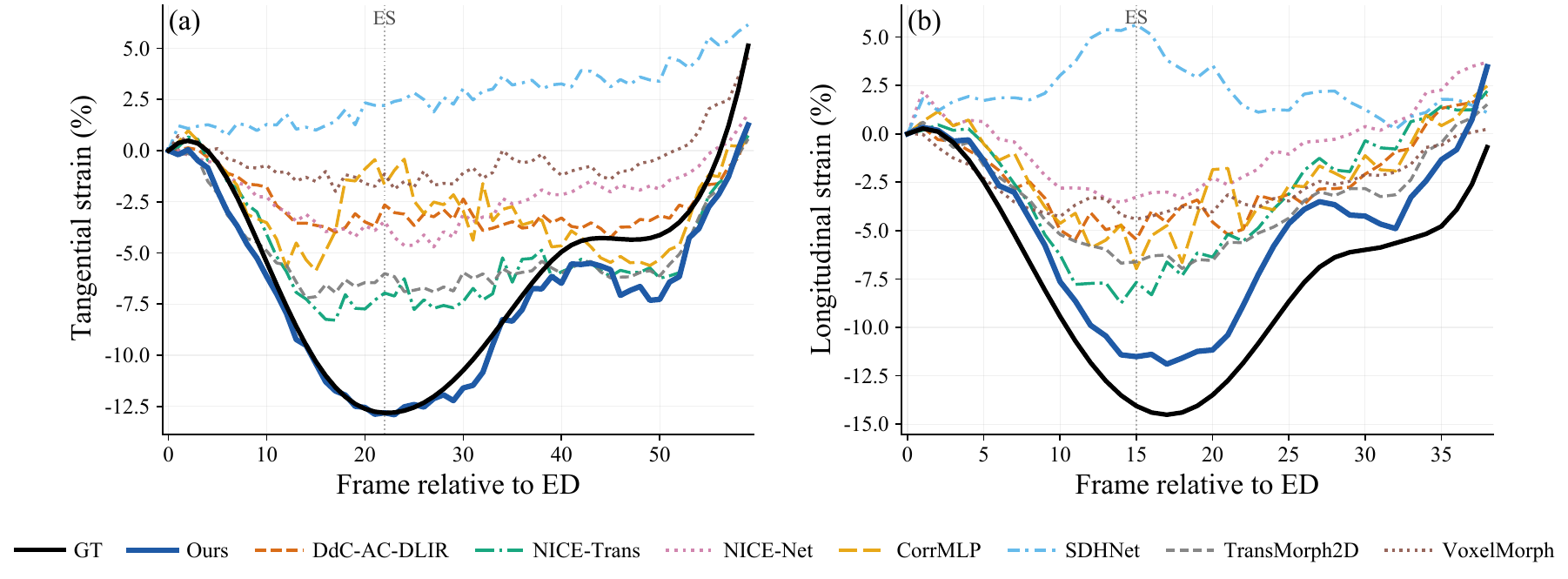}
    \caption{Qualitative comparison of full-cycle global myocardial strain trajectories estimated by different methods on the CAMUS simulation dataset. (a) Global tangential strain and (b) Global longitudinal strain estimated by different methods. The reference trajectory is shown in black, and the ES frame is indicated by a vertical reference line.}
    \label{fig:comparison_global}
\end{figure*}

Fig.~\ref{fig:comparison_global} illustrates the GTS and GLS trajectories estimated by EchoDiST and the comparison methods over the cardiac cycle. EchoDiST closely follows the reference trajectories, capturing both the magnitude and cardiac-phase evolution of myocardial strain. In contrast, the comparison methods show larger deviations and fluctuations from the reference curves over the cardiac cycle.

\begin{table*}[!t]
\centering
\caption{Quantitative comparison of full-cycle global myocardial strain estimation on the CAMUS simulation dataset. All metrics are reported as mean $\pm$ standard deviation over 100 test cases, with global tangential and longitudinal strain curves evaluated over the full cardiac cycle for each case. $p$-values indicate paired statistical comparisons of Curve RMSE between EchoDiST model and each comparison method.}
\label{tab:comparison_global}

{\fontfamily{ptm}\selectfont
\small
\setlength{\tabcolsep}{5pt}

\begin{tabular}{lcccccc}
\toprule
& \multicolumn{3}{c}{Global Tangential Strain}
& \multicolumn{3}{c}{Global Longitudinal Strain} \\
\cmidrule(lr){2-4}
\cmidrule(lr){5-7}

Method
& Curve RMSE (\%)$\downarrow$
& Pearson correlation $r\uparrow$
& $p$-value
& Curve RMSE (\%)$\downarrow$
& Pearson correlation $r\uparrow$
& $p$-value \\
\midrule

DdC-AC-DLIR
& 6.67 $\pm$ 2.68
& \textit{0.666 $\pm$ 0.363}
& $<0.001$
& 8.74 $\pm$ 3.39
& \textit{0.647 $\pm$ 0.361}
& $<0.001$ \\

NICE-Trans
& 5.57 $\pm$ 2.67
& 0.663 $\pm$ 0.360
& $<0.001$
& 8.24 $\pm$ 3.42
& 0.537 $\pm$ 0.422
& $<0.001$ \\

NICE-Net
& 7.10 $\pm$ 2.76
& 0.605 $\pm$ 0.449
& $<0.001$
& 9.76 $\pm$ 3.50
& 0.463 $\pm$ 0.511
& $<0.001$ \\

CorrMLP
& 6.74 $\pm$ 2.95
& 0.441 $\pm$ 0.441
& $<0.001$
& 9.33 $\pm$ 3.55
& 0.364 $\pm$ 0.479
& $<0.001$ \\

SDHNet
& 10.04 $\pm$ 3.38
& $-0.377 \pm 0.442$
& $<0.001$
& 12.66 $\pm$ 4.17
& $-0.407 \pm 0.490$
& $<0.001$ \\

TransMorph2D
& \textit{5.53 $\pm$ 3.00}
& 0.606 $\pm$ 0.476
& $<0.001$
& \textit{7.02 $\pm$ 3.80}
& 0.588 $\pm$ 0.498
& $<0.001$ \\

VoxelMorph
& 7.39 $\pm$ 2.92
& 0.361 $\pm$ 0.579
& $<0.001$
& 9.55 $\pm$ 3.80
& 0.204 $\pm$ 0.591
& $<0.001$ \\

\textbf{Ours}
& \textbf{2.48 $\pm$ 1.57}
& \textbf{0.946 $\pm$ 0.095}
& ref.
& \textbf{2.98 $\pm$ 1.99}
& \textbf{0.956 $\pm$ 0.060}
& ref. \\

\bottomrule
\end{tabular}
}
\end{table*}

\subsubsection{Motion-Derived Functional and Cardiac-Phase Assessment}

\begin{figure}
    \centering
    \includegraphics[width=\columnwidth]{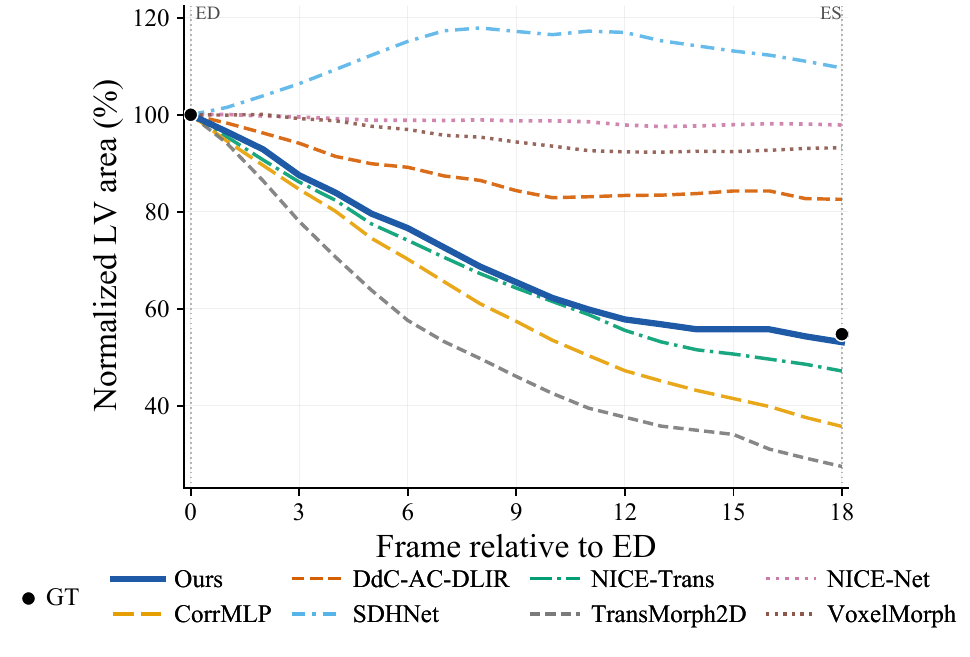}
    \caption{Qualitative comparison of normalized left ventricular area trajectories estimated by different methods on EchoNet-Dynamic. Reference LV-area annotations at ED and ES are shown as discrete black markers.}
    \label{fig:comparison_echonet}
\end{figure}

\begin{table*}[!t]
\centering
\caption{Quantitative comparison of area-based functional accuracy and cardiac-phase consistency on EchoNet-Dynamic. All metrics are evaluated over 110 test cases. Area-based functional metrics are reported at the cohort level, and temporal consistency metrics as mean $\pm$ standard deviation. $p$-values indicate case-level paired statistical comparisons for Area-EF between EchoDiST and each comparison method.}
\label{tab:comparison_echonet}

{\fontfamily{ptm}\selectfont
\small
\setlength{\tabcolsep}{6pt}

\begin{tabular}{lccccc}
\toprule
& \multicolumn{3}{c}{Area-based Functional Accuracy (Area-EF)}
& \multicolumn{2}{c}{Temporal Consistency} \\
\cmidrule(lr){2-4}
\cmidrule(lr){5-6}

Method
& Area-EF RMSE (\%)$\downarrow$
& Area-EF $r\uparrow$
& $p$-value
& ES Localization Error$\downarrow$
& Monotonic Consistency$\uparrow$ \\
\midrule

DdC-AC-DLIR
& 26.90
& 0.383
& $<0.001$
& 2.53 $\pm$ 4.34
& 0.830 $\pm$ 0.211 \\

NICE-Trans
& \textit{13.41}
& 0.355
& 0.011
& 1.17 $\pm$ 0.57
& 0.846 $\pm$ 0.068 \\

NICE-Net
& 33.93
& 0.212
& $<0.001$
& 3.17 $\pm$ 5.54
& 0.768 $\pm$ 0.193 \\

CorrMLP
& 18.57
& \textit{0.472}
& $<0.001$
& \textit{1.16 $\pm$ 0.34}
& \textit{0.858 $\pm$ 0.038} \\

SDHNet
& 38.32
& $-0.209$
& $<0.001$
& 6.13 $\pm$ 7.94
& 0.626 $\pm$ 0.265 \\

TransMorph2D
& 22.59
& 0.406
& $<0.001$
& 1.32 $\pm$ 0.86
& 0.850 $\pm$ 0.076 \\

VoxelMorph
& 33.88
& 0.303
& $<0.001$
& 3.51 $\pm$ 4.77
& 0.746 $\pm$ 0.168 \\

\textbf{Ours}
& \textbf{10.39}
& \textbf{0.686}
& ref.
& \textbf{1.08 $\pm$ 2.85}
& \textbf{0.880 $\pm$ 0.124} \\

\bottomrule
\end{tabular}
}
\end{table*}

Table~\ref{tab:comparison_echonet} presents the motion-derived functional and cardiac-phase assessment of the comparison methods on EchoNet-Dynamic. EchoDiST achieved the lowest Area-EF RMSE and the highest Pearson correlation with the reference Area-EF, indicating more accurate recovery of ED-to-ES chamber contraction. This reduction in Area-EF RMSE was statistically significant compared with all comparison methods (all Holm-adjusted $p<0.05$).

Similar improvements were observed in cardiac-phase assessment, i.e., ED/ES prediction. EchoDiST yielded the lowest ES localization error and the highest monotonic consistency, indicating its potential for cardiac-phase assessment.

Fig.~\ref{fig:comparison_echonet} compares the normalized LV-area trajectories estimated by the comparison methods. EchoDiST shows a more consistent decrease from ED to ES, with an LV-area reduction closer to the reference (black solid markers in Fig.~\ref{fig:comparison_echonet}). In contrast, the comparison methods tend to underestimate the reduction or exhibit larger frame-to-frame fluctuations.

\subsection{Ablation Studies}

To conduct the ablation study, joint anatomy-motion estimation and feedback refinement were selectively enabled or disabled to form four configurations, while all other components and training settings were kept unchanged. For convenience, the corresponding configurations are referred to as w/o Joint \& Feedback, w/o Joint, w/o Feedback, and Full.

\begin{table*}[!t]
\centering
\caption{Quantitative ablation results of anatomical alignment on CAMUS-4CH, CAMUS-2CH, EchoNet-Dynamic, and the CAMUS simulation dataset. All metrics are reported as mean $\pm$ standard deviation over 100, 100, 110, and 100 test cases for the four datasets, respectively, with one ED-to-ES image pair evaluated per case. $p$-values indicate paired statistical comparisons of Dice between the full EchoDiST model and each ablated variant.}
\label{tab:ablation_dice}

{\fontfamily{ptm}\selectfont
\small
\setlength{\tabcolsep}{3.5pt}

\begin{tabular}{cccccccccc}
\toprule
& &
\multicolumn{4}{c}{CAMUS-4CH}
& \multicolumn{4}{c}{CAMUS-2CH} \\
\cmidrule(lr){3-6}
\cmidrule(lr){7-10}

Joint & Feedback
& Dice$\uparrow$
& HD95$\downarrow$
& ASSD$\downarrow$
& $p$-value
& Dice$\uparrow$
& HD95$\downarrow$
& ASSD$\downarrow$
& $p$-value \\
\midrule

&
& 0.888 $\pm$ 0.032
& 7.18 $\pm$ 2.23
& 2.85 $\pm$ 0.89
& $<0.001$
& 0.869 $\pm$ 0.040
& 8.81 $\pm$ 3.01
& 3.54 $\pm$ 1.27
& $<0.001$ \\

& $\checkmark$
& 0.908 $\pm$ 0.024
& 6.10 $\pm$ 2.05
& 2.31 $\pm$ 0.73
& 0.479
& 0.885 $\pm$ 0.030
& 8.26 $\pm$ 2.79
& 3.19 $\pm$ 1.01
& 0.027 \\

$\checkmark$ &
& 0.869 $\pm$ 0.041
& 8.15 $\pm$ 2.86
& 3.36 $\pm$ 1.22
& $<0.001$
& 0.854 $\pm$ 0.043
& 9.47 $\pm$ 3.47
& 4.00 $\pm$ 1.46
& $<0.001$ \\

$\checkmark$ & $\checkmark$
& \textbf{0.909 $\pm$ 0.022}
& \textbf{5.96 $\pm$ 1.80}
& \textbf{2.29 $\pm$ 0.68}
& ref.
& \textbf{0.892 $\pm$ 0.032}
& \textbf{7.47 $\pm$ 2.65}
& \textbf{2.89 $\pm$ 1.03}
& ref. \\

\midrule

& &
\multicolumn{4}{c}{EchoNet-Dynamic}
& \multicolumn{4}{c}{CAMUS Simulation} \\
\cmidrule(lr){3-6}
\cmidrule(lr){7-10}

Joint & Feedback
& Dice$\uparrow$
& HD95$\downarrow$
& ASSD$\downarrow$
& $p$-value
& Dice$\uparrow$
& HD95$\downarrow$
& ASSD$\downarrow$
& $p$-value \\
\midrule

&
& 0.794 $\pm$ 0.069
& 6.70 $\pm$ 2.17
& 3.05 $\pm$ 0.98
& $<0.001$
& 0.748 $\pm$ 0.049
& 7.22 $\pm$ 1.61
& 2.53 $\pm$ 0.50
& $<0.001$ \\

& $\checkmark$
& 0.830 $\pm$ 0.063
& 6.01 $\pm$ 2.03
& 2.51 $\pm$ 0.91
& 0.002
& 0.786 $\pm$ 0.047
& 6.78 $\pm$ 1.81
& 2.16 $\pm$ 0.51
& $<0.001$ \\

$\checkmark$ &
& 0.830 $\pm$ 0.066
& 5.84 $\pm$ 2.24
& 2.57 $\pm$ 1.02
& $<0.001$
& 0.808 $\pm$ 0.043
& 5.43 $\pm$ 1.38
& 1.93 $\pm$ 0.41
& $<0.001$ \\

$\checkmark$ & $\checkmark$
& \textbf{0.888 $\pm$ 0.054}
& \textbf{4.19 $\pm$ 2.04}
& \textbf{1.67 $\pm$ 0.84}
& ref.
& \textbf{0.878 $\pm$ 0.037}
& \textbf{3.58 $\pm$ 1.24}
& \textbf{1.20 $\pm$ 0.33}
& ref. \\

\bottomrule
\end{tabular}
}
\end{table*}

\begin{figure*}
    \centering
    \includegraphics[width=0.9\textwidth,trim=0.8cm 0 0.8cm 0.2cm,clip]{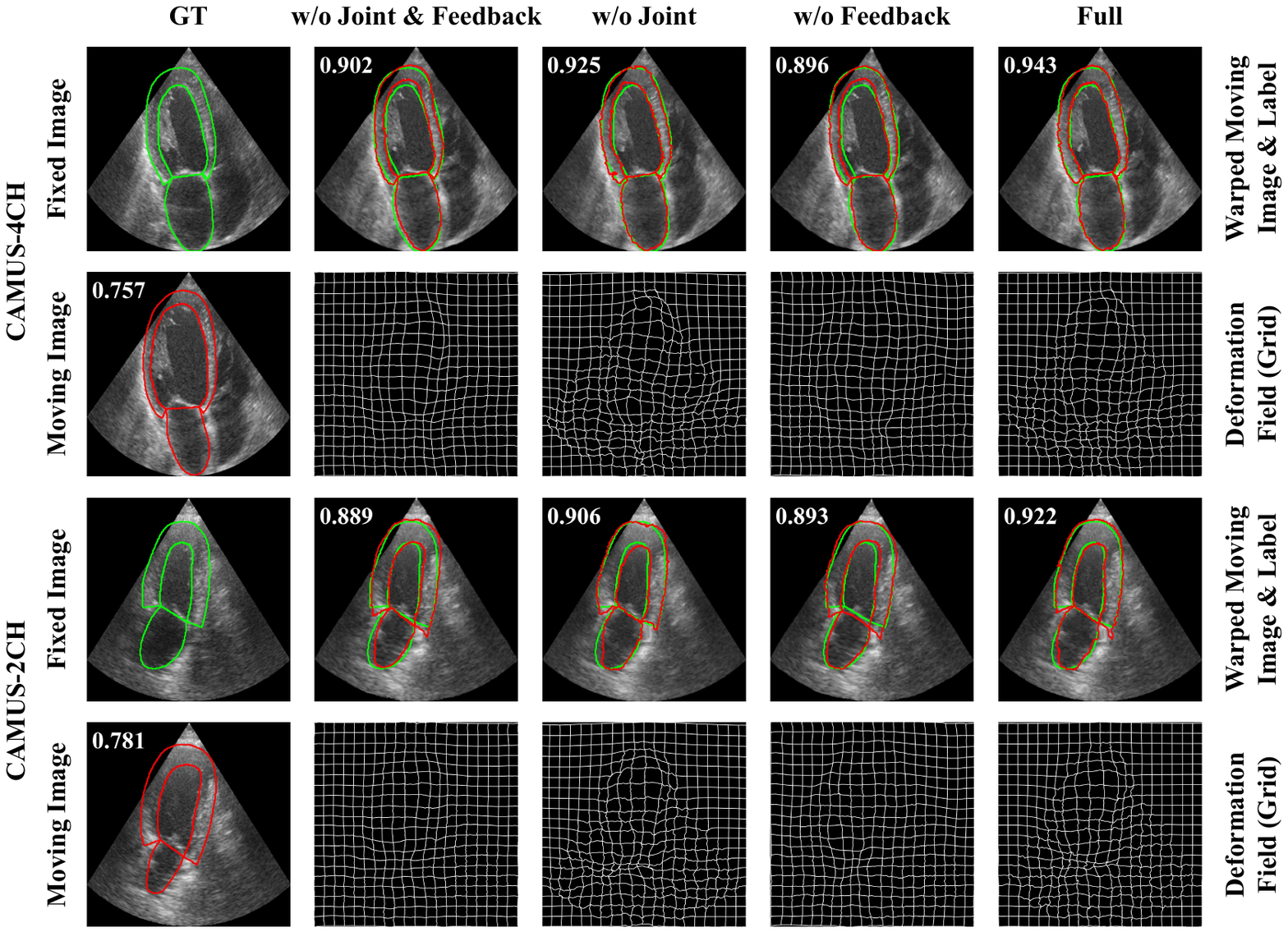}
    \caption{Qualitative comparison of different model variants on the CAMUS dataset. The first column shows the ES fixed image and the ED moving image with anatomical contours, and the remaining columns show the alignment results of different model variants. The dataset corresponding to each two-row group is indicated along the left margin. Within each group, the first row presents the warped moving image with the fixed anatomy in green and the warped moving anatomy in red, while the second row presents the corresponding displacement grid. The numbers displayed on the upper-left corner of each warped-image panel indicate the Dice score for the representative ED-to-ES image pair shown.}
    \label{fig:ablation_dice_1}
\end{figure*}

\begin{figure*}
    \centering
    \includegraphics[width=0.9\textwidth,trim=0.8cm 0 0.8cm 0.2cm,clip]{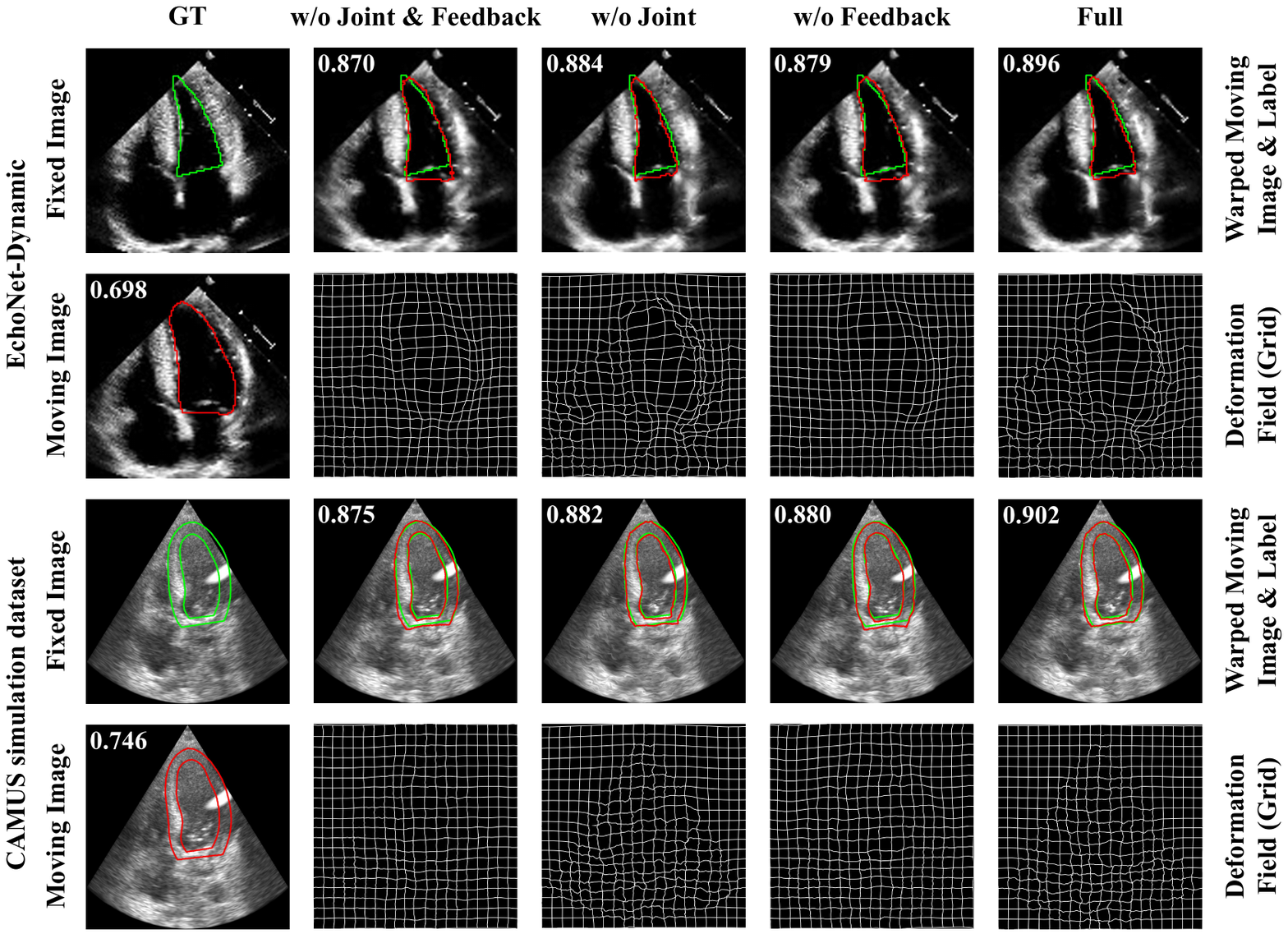}
    \caption{Qualitative comparison of different model variants on EchoNet-Dynamic and the CAMUS simulation dataset. The first column shows the ES fixed image and the ED moving image with anatomical contours, and the remaining columns show the alignment results of different model variants. The dataset corresponding to each two-row group is indicated along the left margin. Within each group, the first row presents the warped moving image with the fixed anatomy in green and the warped moving anatomy in red, while the second row presents the corresponding displacement grid. The numbers displayed on the upper-left corner of each warped-image panel indicate the Dice score for the representative ED-to-ES image pair shown.}
    \label{fig:ablation_dice_2}
\end{figure*}

Table~\ref{tab:ablation_dice} reports the anatomical alignment results across the four datasets. Feedback refinement alone recovered much of the alignment performance, particularly on CAMUS-4CH, whereas joint learning alone showed a less uniform effect across datasets. Combining both components yielded the best performance. The Full model significantly outperformed w/o Joint \& Feedback and w/o Feedback for all three metrics across all datasets. Fig.~\ref{fig:ablation_dice_1} and~\ref{fig:ablation_dice_2} provide further qualitative evidence for the benefit of combining the two components.

\begin{table*}[!t]
\centering
\caption{Quantitative ablation results of myocardial point motion and local strain estimation on the CAMUS simulation dataset. All metrics are reported as mean $\pm$ standard deviation over 100 test cases, with 180 myocardial points evaluated per case. $p$-values indicate paired statistical comparisons of EPE between the full EchoDiST model and each ablated variant.}
\label{tab:ablation_local}

{\fontfamily{ptm}\selectfont
\small
\setlength{\tabcolsep}{8pt}

\begin{tabular}{cccccc}
\toprule
Joint
& Feedback
& EPE (mm)$\downarrow$
& Radial RMSE (\%)$\downarrow$
& Tangential RMSE (\%)$\downarrow$
& \multicolumn{1}{c}{$p$-value} \\
\midrule

&
& 8.10 $\pm$ 2.00
& 19.71 $\pm$ 5.80
& 21.00 $\pm$ 4.80
& $<0.001$ \\

& $\checkmark$
& 7.83 $\pm$ 2.03
& 26.21 $\pm$ 5.40
& 27.16 $\pm$ 5.21
& $<0.001$ \\

$\checkmark$ &
& 6.23 $\pm$ 1.74
& 18.82 $\pm$ 4.42
& 17.69 $\pm$ 3.92
& $<0.001$ \\

$\checkmark$ & $\checkmark$
& \textbf{4.98 $\pm$ 1.58}
& \textbf{16.59 $\pm$ 4.27}
& \textbf{16.50 $\pm$ 5.33}
& ref. \\

\bottomrule
\end{tabular}
}
\end{table*}

\begin{table*}[!t]
\centering
\caption{Quantitative ablation results of full-cycle global myocardial strain estimation on the CAMUS simulation dataset. All metrics are reported as mean $\pm$ standard deviation over 100 test cases, with global tangential and longitudinal strain curves evaluated over the full cardiac cycle for each case. $p$-values indicate paired statistical comparisons of Curve RMSE between the full EchoDiST model and each ablated variant.}
\label{tab:ablation_global}

{\fontfamily{ptm}\selectfont
\small
\setlength{\tabcolsep}{5pt}

\begin{tabular}{cccccccc}
\toprule
& &
\multicolumn{3}{c}{Global Tangential Strain}
& \multicolumn{3}{c}{Global Longitudinal Strain} \\
\cmidrule(lr){3-5}
\cmidrule(lr){6-8}

Joint
& Feedback
& Curve RMSE (\%)$\downarrow$
& Pearson correlation $r\uparrow$
& $p$-value
& Curve RMSE (\%)$\downarrow$
& Pearson correlation $r\uparrow$
& $p$-value \\
\midrule

&
& 8.36 $\pm$ 2.40
& 0.571 $\pm$ 0.419
& $<0.001$
& 10.69 $\pm$ 2.97
& 0.605 $\pm$ 0.387
& $<0.001$ \\

& $\checkmark$
& 8.00 $\pm$ 2.73
& 0.505 $\pm$ 0.481
& $<0.001$
& 10.13 $\pm$ 3.23
& 0.585 $\pm$ 0.409
& $<0.001$ \\

$\checkmark$ &
& 5.71 $\pm$ 2.11
& 0.802 $\pm$ 0.228
& $<0.001$
& 7.27 $\pm$ 2.76
& 0.829 $\pm$ 0.163
& $<0.001$ \\

$\checkmark$ & $\checkmark$
& \textbf{2.48 $\pm$ 1.57}
& \textbf{0.946 $\pm$ 0.095}
& ref.
& \textbf{2.98 $\pm$ 1.99}
& \textbf{0.956 $\pm$ 0.060}
& ref. \\

\bottomrule
\end{tabular}
}
\end{table*}

\begin{figure*}
    \centering
    \includegraphics[width=0.9\textwidth,trim=0.2cm 2.2cm 0 2.3cm,clip]{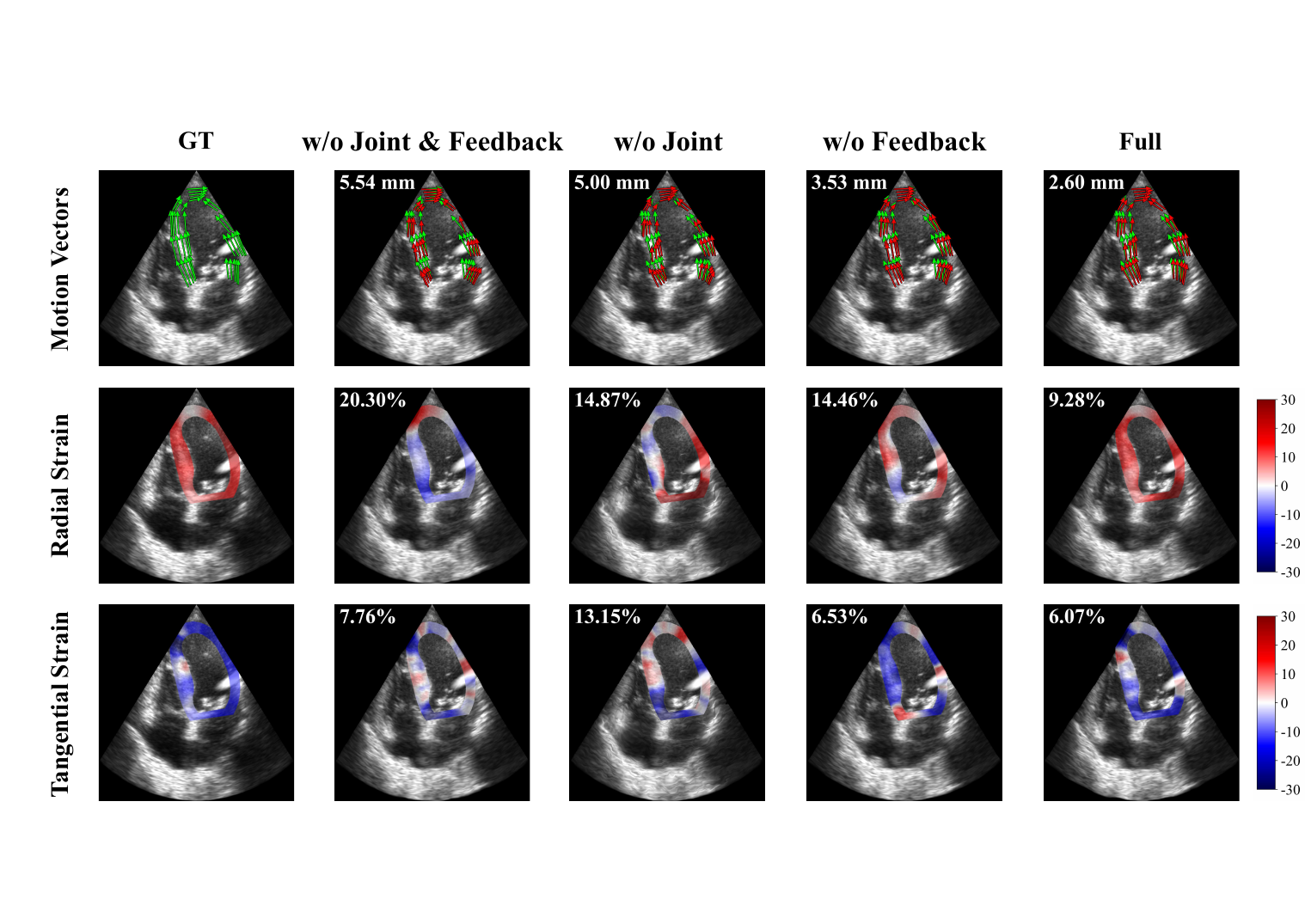}
    \caption{Qualitative comparison of different model variants for myocardial point motion and local strain estimation on the CAMUS simulation dataset. The first row shows ED-to-ES myocardial point-motion vectors, with reference and predicted motions shown in green and red, respectively. The second and third rows show the corresponding radial and tangential strain maps. EPE and strain RMSE are computed over the 180 myocardial points of the representative case and displayed on the upper-left corner of the corresponding motion and strain panels, respectively.}
    \label{fig:ablation_local}
\end{figure*}

\begin{figure*}
    \centering
    \includegraphics[width=0.96\textwidth]{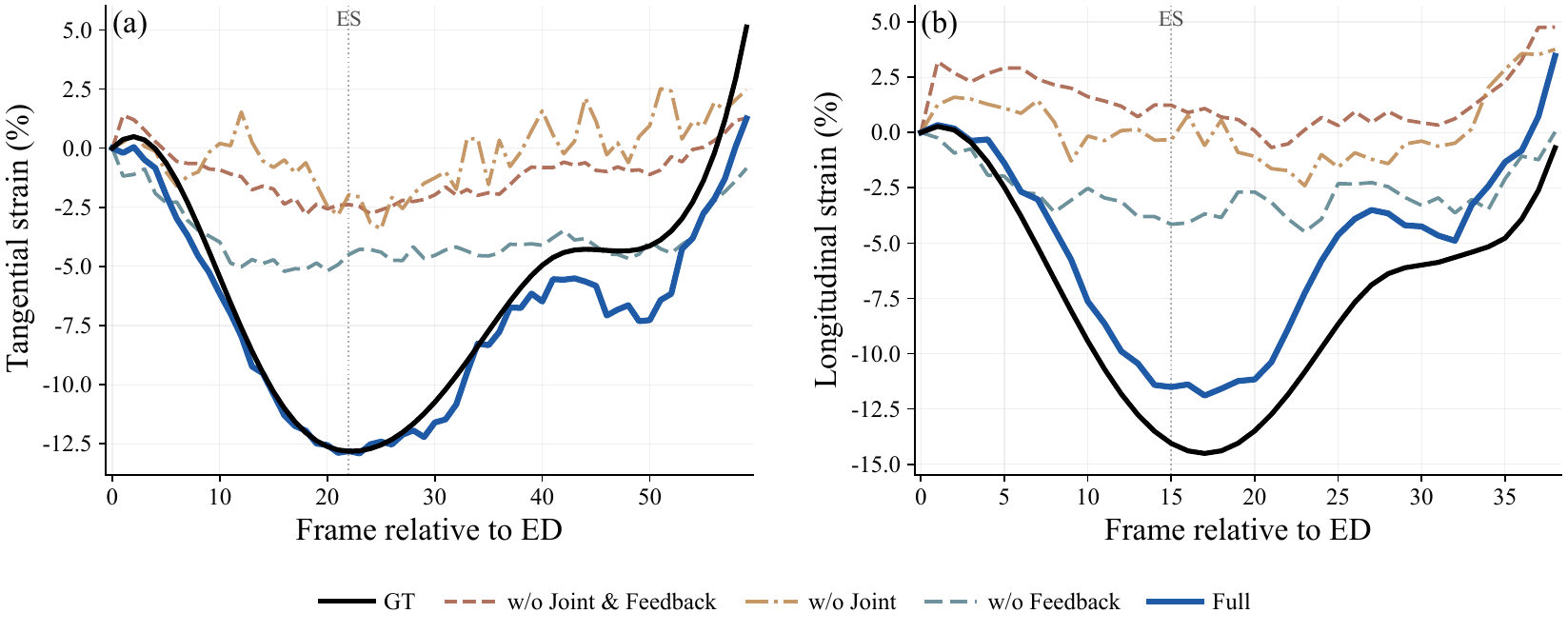}
    \caption{Qualitative comparison of full-cycle global myocardial strain trajectories estimated by different model variants on the CAMUS simulation dataset. (a) Global tangential strain and (b) Global longitudinal strain estimated by different model variants. The reference trajectory is shown in black, and the ES frame is indicated by a vertical reference line. }
    \label{fig:ablation_global}
\end{figure*}

For myocardial motion and strain assessment, the two components showed different effects when evaluated individually. Retaining joint anatomy-motion estimation reduced both point-motion and strain errors, whereas feedback refinement alone slightly reduced point-motion error but did not improve local strain estimation. As summarized in Tables~\ref{tab:ablation_local} and~\ref{tab:ablation_global}, combining the two components yielded the lowest EPE and local strain errors, together with the lowest curve RMSE and the highest correlation for GTS and GLS. The Full model also significantly outperformed all three ablation variants for both global strain measures (all Holm-adjusted $p<0.001$). Fig.~\ref{fig:ablation_local} and~\ref{fig:ablation_global} further support these findings, showing more accurate point motion and more consistent local strain distributions and full-cycle strain trajectories.

\begin{table*}[!t]
\centering
\caption{Quantitative ablation results of area-based functional accuracy and cardiac-phase consistency on EchoNet-Dynamic. All metrics are evaluated over 110 test cases. Area-based functional metrics are reported at the cohort level, and temporal consistency metrics as mean $\pm$ standard deviation. $p$-values indicate case-level paired statistical comparisons for Area-EF between the full EchoDiST model and each ablated variant.}
\label{tab:ablation_echonet}

{\fontfamily{ptm}\selectfont
\small
\setlength{\tabcolsep}{6pt}

\begin{tabular}{ccccccc}
\toprule
& &
\multicolumn{3}{c}{Area-based Functional Accuracy (Area-EF)}
& \multicolumn{2}{c}{Temporal Consistency} \\
\cmidrule(lr){3-5}
\cmidrule(lr){6-7}

Joint
& Feedback
& Area-EF RMSE (\%)$\downarrow$
& Area-EF $r\uparrow$
& $p$-value
& ES Localization Error$\downarrow$
& Monotonic Consistency$\uparrow$ \\
\midrule

&
& 19.43
& 0.620
& $<0.001$
& 1.15 $\pm$ 0.65
& 0.791 $\pm$ 0.034 \\

& $\checkmark$
& 17.42
& 0.683
& $<0.001$
& 1.53 $\pm$ 1.87
& 0.864 $\pm$ 0.064 \\

$\checkmark$ &
& 35.20
& 0.533
& $<0.001$
& 5.86 $\pm$ 8.29
& 0.641 $\pm$ 0.271 \\

$\checkmark$ & $\checkmark$
& \textbf{10.39}
& \textbf{0.686}
& ref.
& \textbf{1.08 $\pm$ 2.85}
& \textbf{0.880 $\pm$ 0.124} \\

\bottomrule
\end{tabular}
}
\end{table*}

\begin{figure}
    \centering
    \includegraphics[width=\columnwidth]{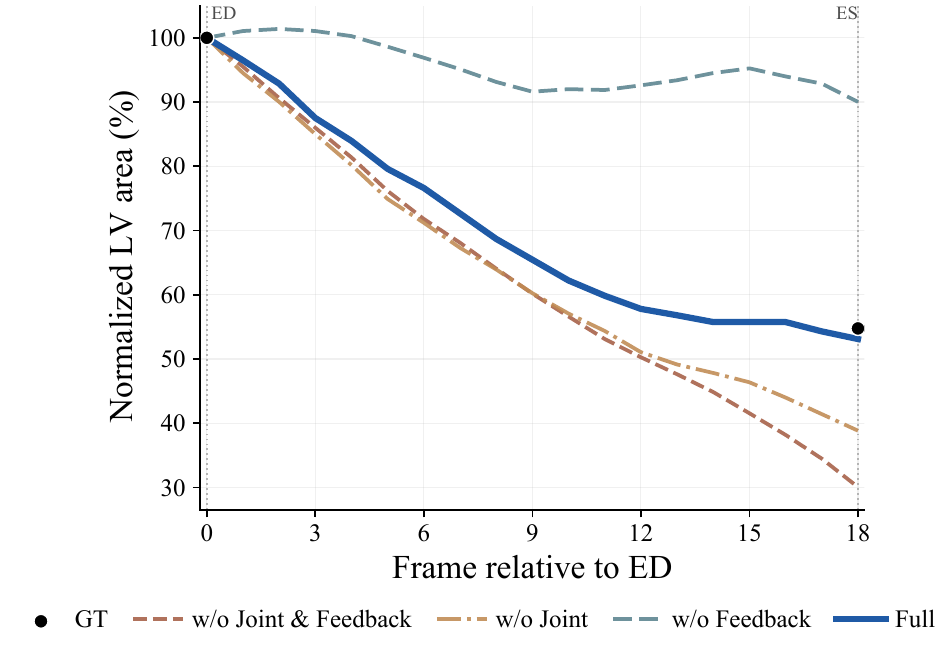}
    \caption{Qualitative comparison of normalized left ventricular area trajectories estimated by different model variants on EchoNet-Dynamic. Reference LV-area annotations at ED and ES are shown as discrete black markers.}
    \label{fig:ablation_echonet}
\end{figure}

This complementary benefit was also evident on EchoNet-Dynamic. With both components enabled, EchoDiST achieved the strongest overall functional and cardiac-phase performance (Table~\ref{tab:ablation_echonet}), with a significant improvement in Area-EF RMSE over all ablation variants. Fig.~\ref{fig:ablation_echonet} further illustrates a smoother and more consistent ED-to-ES LV-area trajectory with fewer frame-to-frame fluctuations.

Overall, the ablation results indicate complementary contributions of joint anatomy-motion estimation and feedback refinement, with their combination providing the most consistent performance across the evaluated tasks.

\section{Discussion}

The main findings of this study are as follows: 1) EchoDiST improves echocardiographic motion estimation by jointly learning anatomical segmentation and motion under limited anatomical supervision; 2) maintains consistent performance across different echocardiographic views and external test datasets, with the improvements extending beyond anatomical alignment to myocardial motion and mechanical assessment; and 3) supports motion-derived functional and cardiac-phase assessment on independently acquired clinical echocardiographic data.

EchoDiST employs a self-distillation-based joint learning mechanism for anatomical segmentation and myocardial motion estimation rather than treating anatomy as an independent auxiliary task. Anatomical segmentation provides structural information about cardiac regions that complements intensity-based correspondence in echocardiographic images. By incorporating this information into motion optimization, cross-frame correspondence is learned from both image intensity and cardiac structure.

The self-distillation-based joint learning mechanism forms a reciprocal interaction. During optimization, the estimated motion propagates anatomical segmentations across frames, while the agreement between the propagated and target anatomy contributes in turn to motion estimation. In this way, anatomical segmentations and motion estimation constrain each other rather than being optimized independently. This interaction may help reduce local inconsistencies in myocardial motion that have limited influence on segmentation accuracy but become more apparent in strain assessment. Feedback refinement serves a different role by correcting residual mismatches during coarse-to-fine motion estimation.

Although this study evaluated the proposed method on three datasets, the available reference data impose important limitations. In the CAMUS simulation dataset, dense point trajectories facilitate controlled evaluation of myocardial motion and strain, but these trajectories are generated by the simulation procedure and should not be interpreted as physiological ground-truth motion. \citet{Evain2022} also noted that local motion may differ from realistic myocardial behavior despite plausible global motion patterns, and that the simulated reverberation model does not capture the full complexity of artifacts encountered in clinical echocardiography. On EchoNet-Dynamic, expert annotations are available only at ED and ES, precluding direct validation of myocardial motion at intermediate frames.

Three additional limitations should be noted. First, the multi-stage training procedure and diffusion-conditioned motion estimation increase the computational cost of model development, although inference requires only a single deterministic forward pass. Second, model development was restricted to CAMUS-4CH. Although evaluation on CAMUS-2CH, the CAMUS simulation dataset, and EchoNet-Dynamic without dataset-specific fine-tuning demonstrates transfer across the tested settings, broader generalization across institutions, scanners, acquisition protocols, image quality, and patient populations remains to be established. In addition, the current framework can only estimate two-dimensional in-plane motion and cannot capture out-of-plane myocardial displacement. Future work will focus on simplifying the training procedure and extending validation to larger multi-center and more heterogeneous echocardiographic cohorts, as well as to multi-view or three-dimensional echocardiography with richer frame-wise clinical motion references.

\section{Conclusion}

This study proposes a diffusion-conditioned framework (EchoDiST) for echocardiographic motion estimation. It combines a diffusion-conditioned motion-estimation network that uses diffusion perturbations as auxiliary conditioning with an EMA-based self-distillation strategy for jointly learning anatomical segmentation and motion estimation. The self-distillation strategy enables joint learning using only limited annotations, allowing the two tasks to be optimized together without extensive manual labeling.

Experimental results across different echocardiographic views and datasets demonstrate consistent improvements in myocardial motion estimation as well as in mechanical, functional, and cardiac-phase assessment. These findings support the potential of EchoDiST for quantitative analysis of cardiac motion and function from echocardiographic images.











\FloatBarrier

\printcredits

\section*{Declaration of competing interest}

The authors declare that they have no known competing financial interests or personal relationships that could have appeared to influence the work reported in this paper.

\section*{Data Availability}

The datasets analyzed in this study are publicly available from their respective repositories. CAMUS, EchoNet-Dynamic, and the CAMUS simulation dataset were used in accordance with their respective data-access requirements. No new patient data were collected for this study. The source code for EchoDiST is not publicly available at the time of submission and will be released upon publication.

\section*{Declaration of generative AI and AI-assisted technologies in the manuscript preparation process}

During the preparation of this work the authors used OpenAI ChatGPT in order to assist with language refinement and manuscript editing. After using this tool, the authors reviewed and edited the content as needed and take full responsibility for the content of the published article.

\section*{Acknowledgments}
This work was supported in part by the National Natural Science Foundation of China (NSFC) under Grants 62531004 and 62027901.

\bibliographystyle{cas-model2-names}

\bibliography{my_EchoDiST}



\end{document}